\documentclass[conference]{IEEEtran}
\IEEEoverridecommandlockouts

\usepackage{times}
\usepackage[T1]{fontenc}

\usepackage{cite}
\usepackage{amsmath,amssymb,amsfonts}
\usepackage{algorithmic}
\usepackage{algorithm}
\usepackage{graphicx}
\usepackage{textcomp}
\usepackage{xcolor}
\usepackage{subcaption}
\usepackage{caption}
\usepackage{booktabs}
\usepackage{tcolorbox}
\usepackage{url}
\usepackage{adjustbox}
\usepackage{enumitem}

\def\BibTeX{{\rm B\kern-.05em{\sc i\kern-.025em b}\kern-.08em
    T\kern-.1667em\lower.7ex\hbox{E}\kern-.125emX}}
\begin{document}

\title{OptiFLIDS: Optimized Federated Learning for Energy-Efficient Intrusion Detection in IoT\\

}

\author{
\IEEEauthorblockN{Saida Elouardi\IEEEauthorrefmark{1}, Mohammed Jouhari\IEEEauthorrefmark{2}, Anas Motii\IEEEauthorrefmark{1}}
\IEEEauthorblockA{\IEEEauthorrefmark{1}College of Computing, University Mohammed VI Polytechnic, Ben Guerir, Morocco \\
Email: \{saida.elouardi, anas.motii\}@um6p.ma}
\IEEEauthorblockA{\IEEEauthorrefmark{2}Laboratory of Research in Informatics (LaRI), Faculty of Sciences, Ibn Tofail University, Kenitra, Morocco \\
Email: mohammed.jouhari1@uit.ac.ma}
}

\maketitle
\begin{abstract}
In critical IoT environments, such as smart homes and industrial systems, effective Intrusion Detection Systems (IDS) are essential for ensuring security. However, developing robust IDS solutions remains a significant challenge. Traditional machine learning-based IDS models typically require large datasets, but data sharing is often limited due to privacy and security concerns. Federated Learning (FL) presents a promising alternative by enabling collaborative model training without sharing raw data. Despite its advantages, FL still faces key challenges, such as data heterogeneity (non-IID data) and high energy and computation costs, particularly for resource-constrained IoT devices. To address these issues, this paper proposes OptiFLIDS, a novel approach that applies pruning techniques during local training to reduce model complexity and energy consumption. It also incorporates a customized aggregation method to better handle pruned models that differ due to non-IID data distributions.  Experiments conducted on three recent IoT IDS datasets, TON\_IoT, X-IIoTID, and IDS-IoT2024, demonstrate that OptiFLIDS maintains strong detection performance while improving energy efficiency, making it well-suited for deployment in real-world IoT environments.
\end{abstract}
\begin{IEEEkeywords}
 Intrusion Detection System, Deep Learning, IoT, Federated Learning, Privacy and Security, Energy Efficiency, Pruning, Non-Independent and Identically Distributed Data
\end{IEEEkeywords}

\section{Introduction}
The Internet of Things (IoT) is increasingly integrated into everyday life, interconnecting devices ranging from household appliances to industrial control systems \cite{Kuzlu2021}. While this pervasive connectivity offers significant benefits, it also expands the attack surface, exposing networks to a growing array of security threats \cite{DESOUZA2022}. To mitigate these risks, multi-layered defense strategies are essential, with Intrusion Detection Systems (IDS) \cite{BENADDI2025101624} playing a central role by continuously monitoring network traffic and detecting both known and emerging attacks.

Machine Learning (ML) and Deep Learning (DL) have significantly improved Intrusion Detection Systems (IDS) by enabling automatic feature extraction from complex network traffic patterns \cite{10767686}. However, centralized training of such models is often impractical due to privacy and regulatory restrictions on data sharing. Federated Learning (FL) addresses this challenge by enabling multiple devices to collaboratively train a shared global model without exposing raw data. Despite these advantages, FL introduces technical challenges, particularly non-IID data distributions and computational constraints, that can degrade convergence and performance.

While deep learning DL-based IDS  can achieve high detection accuracy, their heavy computational and memory demands limit deployment on resource-constrained IoT devices. Pruning has emerged as an effective solution to reduce model size and computational load by removing non-essential parameters while preserving accuracy \cite{Fang2023}. However, applying pruning within FL raises several unresolved challenges. Heterogeneous client datasets can lead to structurally different pruned models, making aggregation less effective. In addition, pruning strategies must balance accuracy preservation with energy efficiency, particularly in non-IID environments.

These limitations motivate the need for an FL-based IDS framework that jointly addresses computational efficiency, energy consumption, and detection performance under realistic IoT conditions. This paper introduces \textit{OptiFLIDS}, a pruning-enhanced FL framework that integrates model compression directly into the federated training process. OptiFLIDS adapts model size to client constraints, employs an aggregation strategy tailored to heterogeneous pruned models, and formulates pruning as a multi-objective optimization problem guided by a Deep Reinforcement Learning (DRL) agent \cite{articledeep}. The key contributions of this work are summarized as follows: 
\begin{enumerate}
    \item We propose OptiFLIDS, a novel pruning methodology designed to enhance both model compression and energy efficiency in FL based on Convolutional Neural Network (CNN) pruning in an IoT environment. We adapt a specialized aggregation method to handle heterogeneous pruned models under non-IID data distributions.
    \item We formulate our pruning approach as a multi-objective optimization problem, aiming to minimize energy consumption while maintaining high performance across all clients. To this end, we employ a DRL to optimize our pruning.
    \item We conduct extensive experiments on recent real-world IoT datasets for multiclass classification under both IID and non-IID settings, using two aggregation algorithms: FedAvg~\cite{aouedi2024survey}  and FedProx~\cite{QI2024272}. To support reproducible research, our complete code is available at this link \footnote{\url{https://github.com/SAIDAELOUARDI23/OptiFLIDS-.git}}.
\end{enumerate}

\begin{figure}[t!]
 \centering
  \includegraphics[width=0.90\columnwidth]{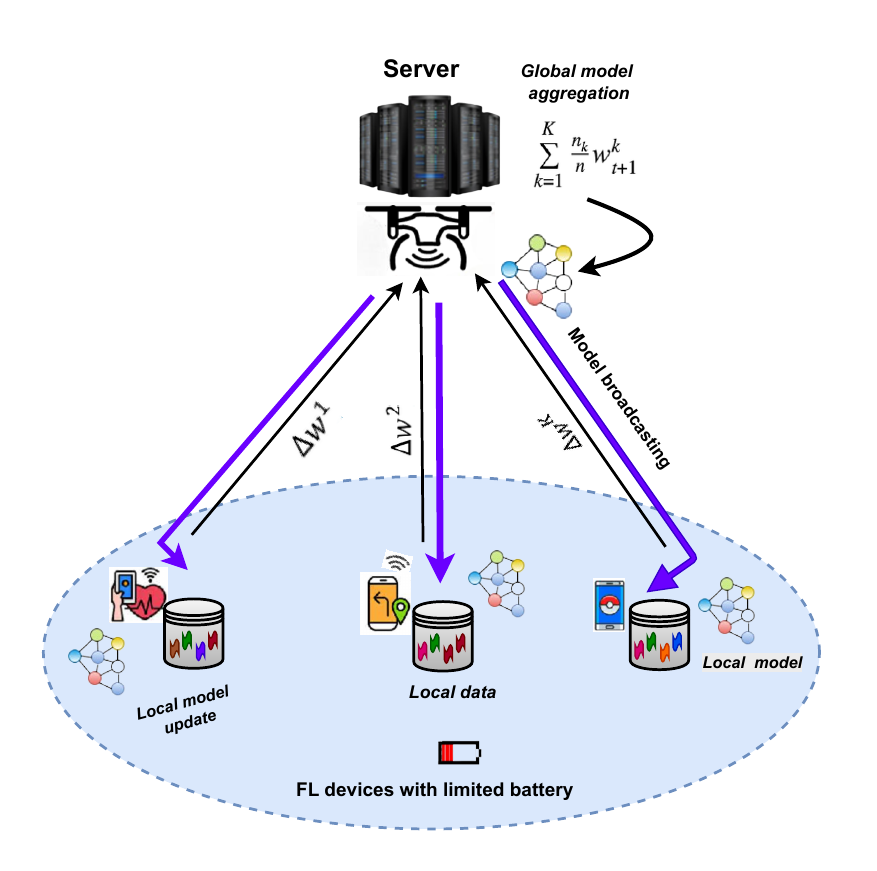}
    \caption{Overview of federated intrusion detection in Resource-constrained IoT environments}
    \label{fig:overview}
   
\end{figure}

To better illustrate the context  addressed in this work, Fig.\ref{fig:overview} presents a general overview of FL-based intrusion detection in resource-constrained IoT environments.

The remainder of this paper is organized as follows: Section~\ref{A} reviews related work on FL-based IDS. Section~\ref{B} presents the OptiFLIDS framework and explains how it addresses the core research problem. Section~\ref{C} reports on the experimental evaluation conducted on three recent IoT datasets. Finally, Section~\ref{D} concludes the paper and discusses the limitations of the approach.

\begin{table*}[ht]
\scriptsize
\renewcommand{\arraystretch}{0.8}
\caption{Existing federated learning approaches for IoT intrusion detection}
\label{tab:fl_iot_ids}
\centering
\begin{tabular}{p{0.6cm} p{1.38cm} p{2.3cm} p{2.2cm} p{4.6cm} p{4.1cm}}
\toprule
\textbf{Ref} & \textbf{Network Type} & \textbf{Technique} & \textbf{Dataset} & \textbf{Main Contributions} & \textbf{Limitations} \\
\midrule
\cite{article21} & Enterprise \& IoT Networks & FL for CTI sharing & NF-UNSW-NB15-v2, NF-BoT-IoT-v2 & Privacy-preserving CTI sharing; improved detection across heterogeneous data sources & Sensitive to non-IID data; moderate computational and communication overhead \\
\midrule
\cite{9416805} & IIoT & FL with blockchain integration & Simulated Turbofan & Demonstrated scalability benefits with increased clients & Limited to positioning attacks; high computational and energy cost \\
\midrule
\cite{9352785} & IoT Edge Computing & Traditional neural network & BoT-IoT & Addressed multiple cyber threats (data exfiltration, keylogging, server scanning, DoS/DDoS over HTTP/TCP/UDP); achieved 99\% AUC & Energy consumption not evaluated \\
\midrule
\cite{9416835} & Vehicular Fog (IIoT) & FL with Laplace and homomorphic encryption & Not specified & Ensures robustness against system variation & High communication and computation cost \\
\midrule
\cite{10747349} & IoT & Structured pruning integrated with FL for NIDS & UNSW-NB15, USTC-TFC2016, CIC-IDS-2017 & Efficient, privacy-preserving intrusion detection with low computation and communication costs & May face challenges with non-IID data, heterogeneous clients, and large-scale deployment \\
\midrule
\cite{9454328} & IIoT & Iterative GRU-based FL model & UNSW-NB15 & Achieved 98\% detection accuracy & Energy impact not assessed; limited to specific DDoS scenarios \\
\midrule
 \cite{9145588} & IIoT & FL + Gradient Sparsification + Quantization + Differential Privacy + Adaptive Scheduling & Custom simulated dataset &  Communication-efficient, local optimization, secure aggregation, trust-aware scheduling, and proactive anomaly detection for reliable IIoT intelligence and low bandwidth use & Limited to simulation; lacks validation on real-world industrial datasets \\
\bottomrule
\end{tabular}
\end{table*}

\section{RELATED WORK} \label{A}

This section reviews recent advances in FL for anomaly detection in IoT environments. We organize the related work into two subsections: FL for IDS, and FL-based IDS combined with compression techniques for enhanced efficiency. Table~\ref{tab:fl_iot_ids} provides a comparative overview of existing FL-based approaches for IoT intrusion detection, highlighting their network types, techniques, datasets, key contributions, and limitations.

\subsection{Federated Learning Approaches for Intrusion Detection}

The work in \cite{article21} proposes a  FL framework designed to overcome challenges related to data privacy and heterogeneity across multiple organizations. Their approach enables collaborative training of a robust IDS model without exchanging sensitive  data. Utilizing uniformly formatted NetFlow datasets, NF-UNSW-NB15-v2 and NF-BoT-IoT-v2. they evaluated the framework under centralized, local, and federated training scenarios. Results indicate that the FL model effectively detects malicious network traffic while maintaining data privacy.

The authors in \cite{9416805} introduced TrustFed, a blockchain-based framework aimed at ensuring fairness and trustworthiness in cross-device FL systems within IIoT environments. TrustFed tackles challenges such as model poisoning by detecting and excluding malicious participants, while leveraging blockchain smart contracts to maintain a reputation system that incentivizes honest contributions.  Experimental results on large-scale IIoT datasets show that TrustFed outperforms conventional methods in terms of fairness, security, and  performance.

A lightweight cyberattack detection framework for IoT edge computing environments is presented in~\cite{9352785}. The framework integrates a multi-attack detection mechanism operating directly at the edge layer, enabling rapid response and reducing cloud workload. It supports both centralized and FL modes, allowing flexible deployment based on system requirements. Evaluation on the BoT-IoT dataset shows superior accuracy and computational efficiency compared to conventional ML and DL methods.

The authors in~\cite{9416835} present a privacy-preserving aggregation mechanism for FL-based navigation in vehicular fog computing. Vehicles collaboratively train models without sharing raw data, and a secure aggregation protocol handles encrypted updates to address bandwidth, dynamic topology, and real-time constraints. Experiments show strong performance with low communication overhead. Similarly,~\cite{9454328} proposes a FL architecture with fog/edge computing for IIoT, allowing nodes to train an IDS collaboratively while preserving privacy. The framework leverages edge intelligence and a collaborative mitigation layer, achieving fast, cost-efficient, and accurate threat detection with 98\% accuracy.

\subsection{Compression-Enhanced Federated Learning for IDS}

A lightweight FL approach for efficient network intrusion detection in resource-constrained environments is proposed in~\cite{10747349}. This method allows multiple devices to collaboratively train a privacy-preserving detection model with minimal computation and communication costs. Experimental results show improved detection accuracy and efficiency, making it suitable for real-world deployment in distributed networks. In \cite{9145588}, the authors propose a Communication-Efficient Federated Learning (CEFL) framework specifically designed for  IIoT environments. To tackle the challenges of limited bandwidth, device heterogeneity, and privacy concerns, CEFL integrates gradient sparsification and quantization techniques to reduce communication costs, while employing differential privacy to ensure secure aggregation of updates. Furthermore, a threat detection module is incorporated to monitor gradient variations and proactively exclude compromised devices. CEFL also includes an adaptive scheduling mechanism that balances device contributions based on energy availability, trust scores, and network conditions. Experimental results in simulated IIoT environments demonstrate that CEFL achieves strong detection performance, reduced communication overhead, and resilience against unreliable or malicious devices, making it a robust solution for large-scale industrial intelligence systems.
\section{PROPOSED SYSTEM MODEL}
 \label{B}

In this section, we discuss the modeling Non-IID data, provide an overview of our proposed FL-based IDS model, and formulate the optimization problem.

\subsection{Modeling Non-IID Data  Using Gamma-Based Partitioning}

Non-IID data distribution is a frequent challenge in FL-based IDS, where clients possess diverse data patterns, attack types, or volumes. This heterogeneity can negatively affect model training, convergence, and  performance. To enhance the robustness of intrusion detection models, it is crucial to address these non-IID challenges. Non-IID data in FL can be classified into three types: quantity skew, where all clients share the same categories but differ in sample sizes; label skew, where clients have data for only some categories; and mixed skew, which combines both quantity and label skews.

To simulate Non-IID data distributions across clients, we use the Gamma distribution with normalization \cite{articlegaama}, a continuous distribution commonly applied in statistics and probability theory. It is defined by two parameters: the shape parameter ($\alpha$) and the scale parameter ($\beta$). Its probability density function (PDF) is given by:
\begin{equation}
f(x; \alpha, \beta) = \frac{x^{\alpha - 1} e^{-\frac{x}{\beta}}}{\beta^\alpha \Gamma(\alpha)}
\end{equation}

where $\Gamma(\alpha)$ represents the Gamma function, expressed as:
\begin{equation}
\Gamma(\alpha) = \int_{0}^{+\infty} x^{\alpha - 1} e^{-x} \, dx
\end{equation}

The shape parameter $\alpha$ of the Gamma distribution critically influences data heterogeneity among clients. When $\alpha$ is small, the distribution is highly skewed, with most of the probability mass near zero, causing a severe imbalance and strong Non-IID effects. In contrast, larger $\alpha$ values produce a more symmetric, Gaussian-like distribution, leading to a more balanced data allocation that approximates IID conditions. 
To partition data among clients, we sample proportions from the Gamma distribution and normalize them to ensure they sum to 1:
\begin{equation}
\theta_i = \frac{\text{distributions}_i}{\sum_{j=1}^{\text{num\_clients}} \text{distributions}_j}
\end{equation}

where $\theta = [\theta_1, ..., \theta_k]$ represents the normalized proportions for each client. These proportions are used to allocate data samples, ensuring a controlled non-IID distribution.

Although the Dirichlet distribution \cite{10400794} is widely used for generating such probability vectors, it samples all components jointly in a single step. In contrast, our approach using  normalized  Gamma allows each class proportion to be generated independently. This enables greater flexibility, such as injecting noise into specific components, applying class-specific biases, or controlling the heterogeneity per client before normalization. Both methods are mathematically equivalent when the Gamma variables are drawn with shape parameters corresponding to the Dirichlet concentration parameters.

\begin{figure}[ht]
    \centering
    \begin{subfigure}[b]{0.36\textwidth}
        \centering
        \includegraphics[width=\textwidth]{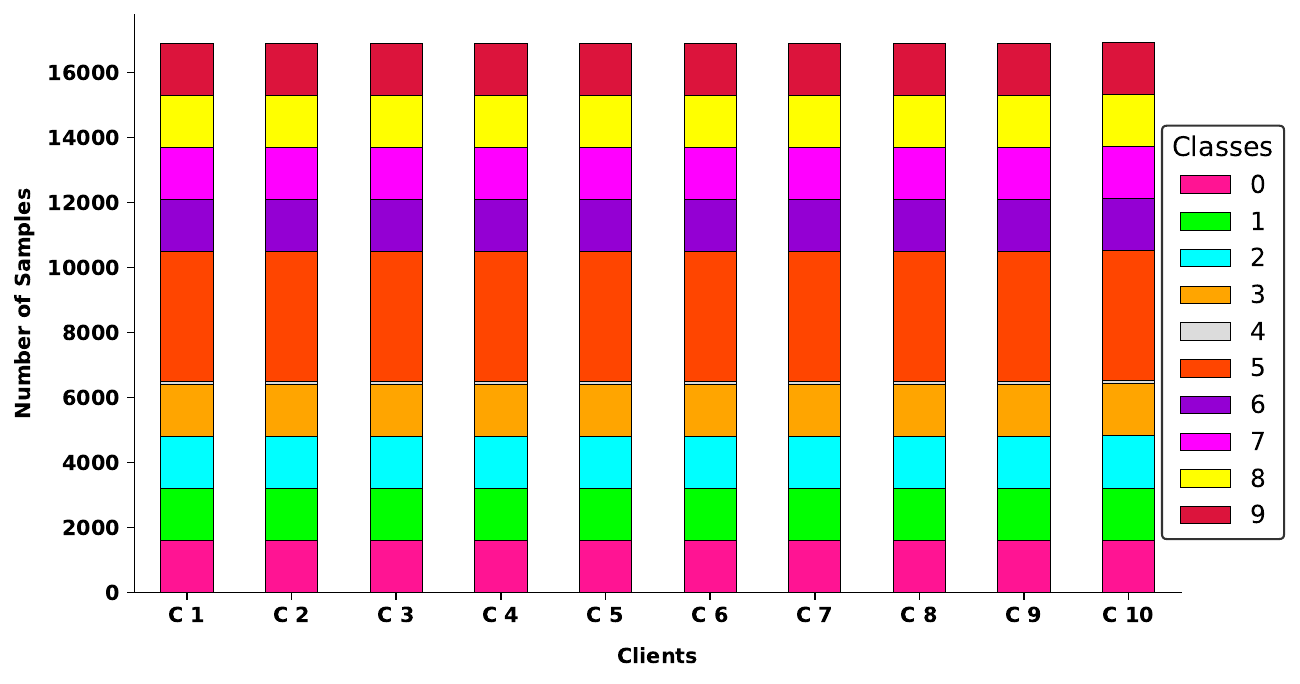}
        \caption{ $\boldsymbol{\alpha} = 1000000$ }
        \label{fig:class_distribution_IID}
    \end{subfigure}
    \hfill
    \begin{subfigure}[b]{0.36\textwidth}
        \centering
        \includegraphics[width=\textwidth]{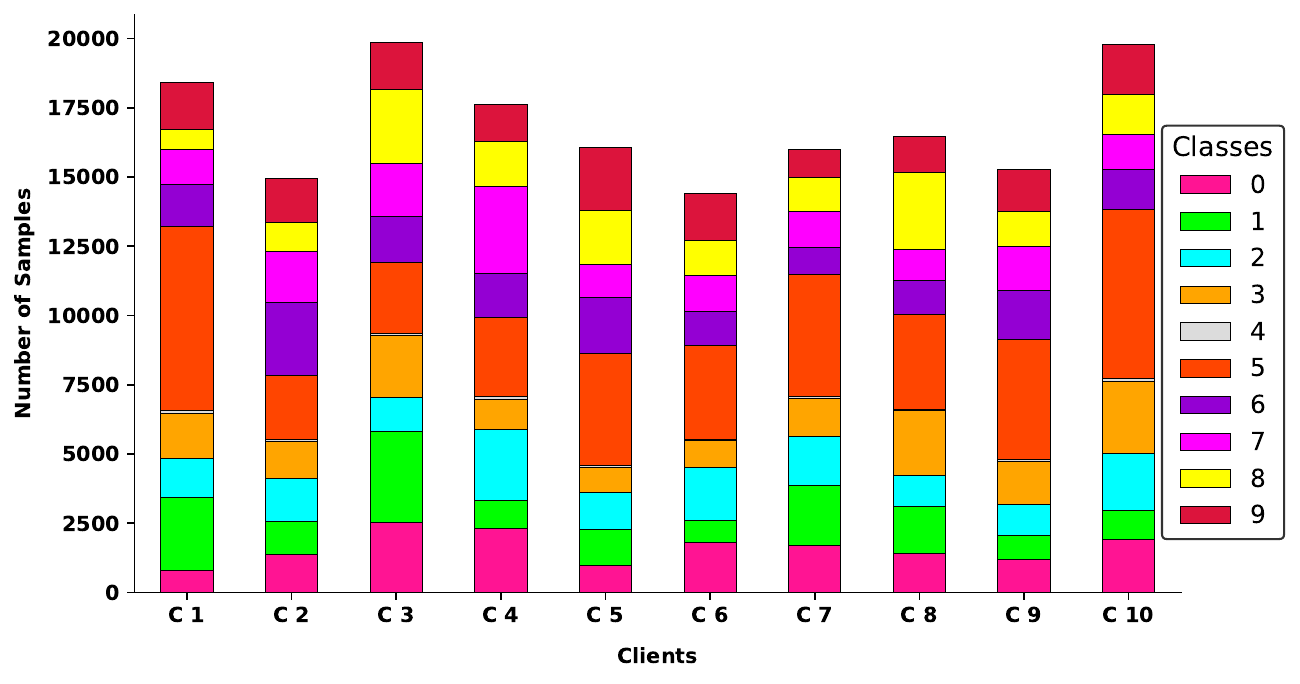}
        \caption{  $\boldsymbol{\alpha} = 10$}
        \label{fig:class_distribution_NonIID}
    \end{subfigure}
    \caption{  Heatmap of data distribution across clients in the Ton\_IoT Dataset Simulated Using Normalized Gamma Distribution with Varying Shape Parameters $\boldsymbol{\alpha}$.}
    \label{fig:class_comparison}
\end{figure}

As shown in Fig.~\ref{fig:class_comparison}, we present the data distribution of 10 attack classes: 'backdoor' (0), 'ddos' (1), 'dos' (2), 'injection' (3), 'mitm' (4), 'normal' (5), 'password' (6), 'ransomware' (7), 'scanning' (8), and 'xss' (9), across 10 clients using the Ton\_IoT dataset \cite{article122} for $\alpha = 10$ and $\alpha = 1,000,000$. With $\alpha = 10$, the distribution is non-IID, showing significant variation across clients. When $\alpha = 1,000,000$, it becomes IID, with more uniform distribution.

 \subsection{ Model Pruning}

In wirelessFL, updating models on resource-constrained devices and communicating them across the network results in significant computational and communication energy costs. To mitigate these overheads, model pruning is commonly employed to reduce model size by eliminating redundant neural connections, without extra requirements. This makes pruning particularly suitable for federated IoT environments. In contrast, quantization techniques face several limitations: dynamic or mixed-precision quantization often necessitates retraining or hardware-specific support, while post-training quantization (PTQ) methods provide only fixed bitwidth models, leading to increased storage requirements and switching overhead in response to varying IoT device capabilities \cite{11049020}. Furthermore, knowledge distillation introduces additional computational load by training a secondary model (student), which is impractical for highly resource-limited IoT devices \cite{article2025}.

Pruning is particularly advantageous  in our context, as it increases model sparsity and reduces both computational and communication costs, which are critical factors in resource-constrained wireless FL. Furthermore, it can be easily integrated into an existing federated system, requiring only local weight adjustments. This simplicity makes pruning a suitable choice for our federated IoT environment.

Pruning insignificant neurons or weights effectively reduces the model size with minimal performance degradation. The learning accuracy only significantly decreases when the pruning ratio is high. According to \cite{Molchanov2019}, the importance of a weight is determined by the error introduced when it is removed, where the induced error is measured as the squared difference in prediction errors with and without the \( j \)-th weight \( w_{k,j} \) of the \( k \)-th device. This is denoted as:  
\begin{equation}
I_{k,j} = \left( F_k(W_k) - F_k(W_k \mid w_{k,j} = 0) \right)^2
\label{eq:importance_sw}
\end{equation}
Where \( F_k(W_k) \) represents the local loss function, and \( W_k \) denotes the local neural network parameters of the \( k \)-th  device. A larger error indicates a higher weight significance. However, computing \( I_{k,j} \) for each weight in the \( k \)-th edge device, as defined in (\ref{eq:importance_sw}), is computationally intensive, particularly when the model contains a large number of weights.  

To reduce the computational cost of importance estimation, we approximate it by measuring the magnitude of the \( j \)-th updated weight \( \hat{w}_{k,j} \) of the local model of the \( k \)-th device as follows:
\begin{equation}
\hat{I}_{k,j} = \left| \hat{w}_{k,j} \right|,
\label{eq:L1}
\end{equation}
\subsection{The Proposed FL  Framework}
The proposed OptiFLIDS framework with pruning is illustrated in Fig. \ref{fig:model_overview1}, where the weights of the local models are updated and pruned on the local  devices, and then aggregated on the  server using a specialized aggregation method. The learning process is updated as follows:
\begin{figure}[ht]
  \includegraphics[width=1.03\columnwidth]{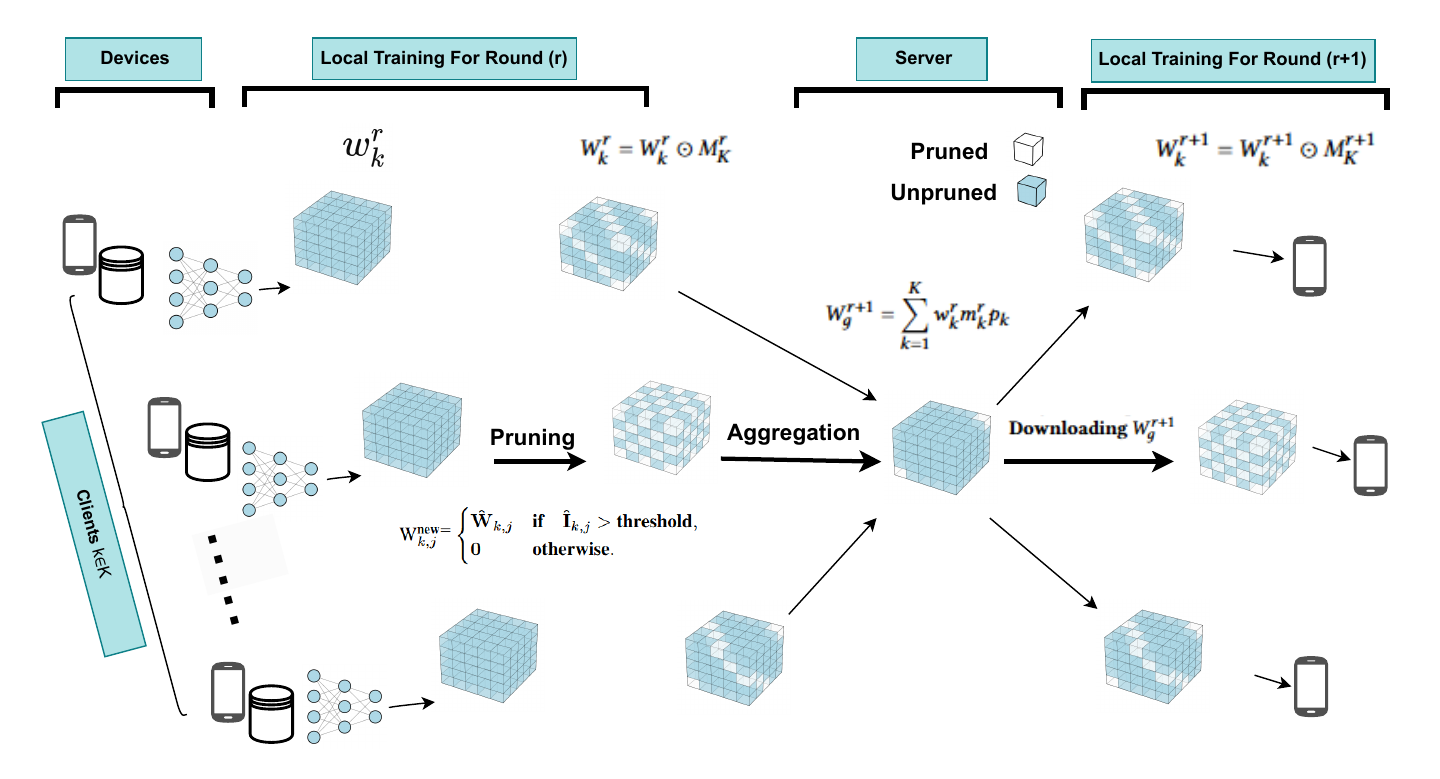}
    \caption{ Overview of the proposed approach for our federated pruning during the first round of federated learning}
    \label{fig:model_overview1}
\end{figure}

\begin{itemize}[left=0pt]
\item Global Model Broadcasting: During the first round (\( q = 0 \)), the server initializes the global model \( \theta_g \) and sends \( \theta_g \) to all devices via downlink communication.

\item Local Model Updating: Each client \( k \) initializes the local model with the global parameters, setting \( \theta_k = \theta_g \), and then updates the local model using the Adam optimizer \cite{hospodarskyy2024understanding} with learning rate \( \eta_k \). The update at iteration \( t \) of round \( q \) is given by :
\begin{equation}
W_k^{q,t+1} = W_k^{q,t} - \eta_k \frac{\hat{m}_k^{q,t}}{\sqrt{\hat{v}_k^{q,t}} + \epsilon}
\end{equation}
where \( \hat{m}_k^{q,t} \) and \( \hat{v}_k^{q,t} \) are bias-corrected first and second moment estimates of the gradients.\\[0.1cm]
    After training the local model for \( E \) epochs, the pruning process begins by calculating the importance of each weight in the local model, as illustrated in Fig.~\ref{fig:weights pruning}. Based on these importance scores, the weights are sorted in descending order. A pruning ratio is then defined to determine the percentage of the least important weights to be removed.
\begin{figure}[ht]
    \includegraphics[width=1.03\columnwidth]{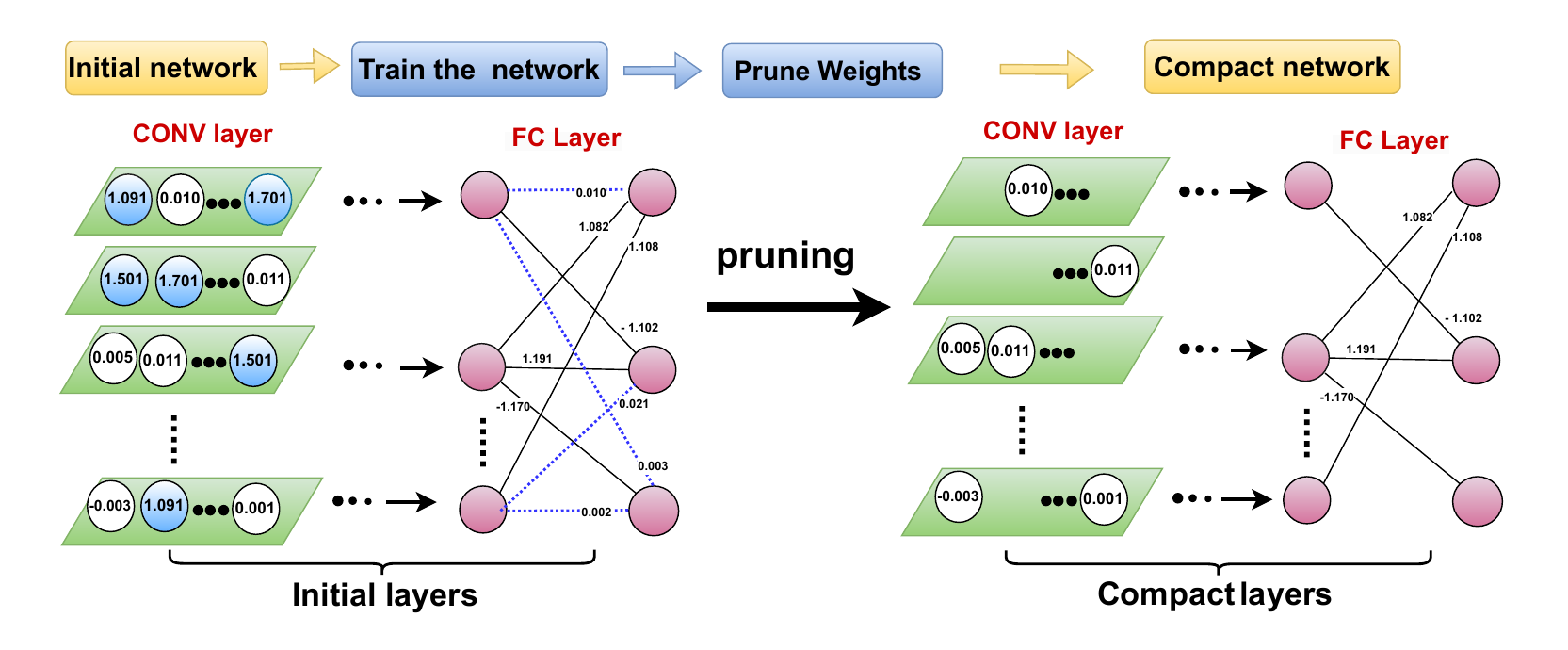}
    \caption{Flowchart of weights pruning procedure}
    \label{fig:weights pruning} 
\end{figure}

The pruning operation is performed only during the first communication round, as this round is sufficient to assess the importance of the updated weights after training. A pruning mask is then generated, matching the size and structure of the model’s weights. This mask assigns a value of 1 to the top-ranked weights (to keep) and 0 to the lowest-ranked weights (to prune). The mask is applied directly to the weights, ensuring that only the important weights remain active, while the pruned weights are set to zero for all subsequent rounds.
The pruning operation is applied as follows:
\begin{equation}
W_k = W_k \odot M_k
\end{equation}

where \( M_k \) is the pruning mask. After pruning, a fine-tuning process is performed to recover any potential loss in performance. Given a pruning ratio \( \rho_k \) for the \( k \)th device, the number of remaining weights after pruning is calculated as: 
\begin{equation}
W_{\rho_k} =  (1 - \rho_k) W_k
\end{equation}

After local training, each client sends its updated model weights to the server. The pruning mask \( M_k \) is also transmitted, but only during the first communication round. The server stores this mask and reuses it for all subsequent aggregation operations. This strategy not only ensures consistent weight selection across rounds, but also significantly reduces communication overhead by avoiding repeated transmission of the pruning mask.

    \item Model Aggregation:  In FL, each client has a different local dataset, which is often non-IID . As a result, the models trained by different clients may differ, even if the pruning rate is the same. For example, a weight might be pruned in client \( k \), but not in client \( j \). This creates differences in the model structures, which makes it difficult to update the global model effectively. To handle this, our OptiFLIDS framework performs specialized aggregation only on the weights that are not pruned across clients. These unpruned weights are considered important, as they are shared among clients and reflect useful informations from different datasets, as illustrated in Fig.~\ref{fig:agg}.

\begin{figure}[ht]
  \centering
   \includegraphics[width=0.8\columnwidth]{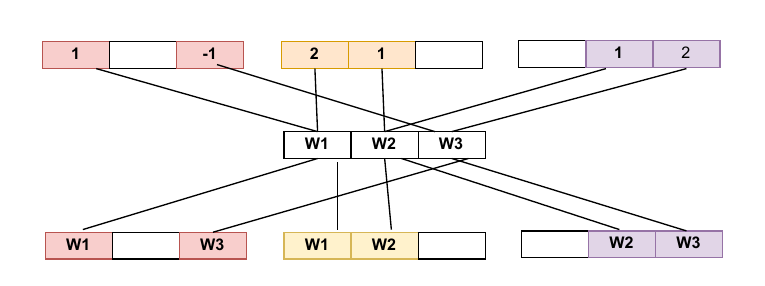}
    \caption{Illustration of aggregation in OptiFLIDS: only unpruned weights are aggregated across the three clients}
    \label{fig:agg} 
\end{figure}

The aggregation of the shared, unpruned weights is defined as:
\begin{equation}
W_{\text{global}}^{q+1} = \sum_{k=1}^{K} \sum_{j=1}^{P} w_{k,j}^{q}  p_k  m_{k,j}
\end{equation}
The weighting factor \( p_k \) for client \( k \) can be defined based on the size of their local dataset \( D_k \) as follows:
\begin{equation}
p_k = \frac{|D_k|}{\sum_{i=1}^{P} |D_i|},
\end{equation}

Where \( |D_k| \) represents the size of the local dataset for client \( k \), and \( \sum_{i=1}^{P} |D_i| \) is the sum of the sizes of the datasets for all \( K \) clients.

The aggregated weights are sent back to each participating client \( k \), where the unpruned weights are updated for the  \( q+1 \) communication round. This strategy allows the server to retain the most relevant weights for each client. The local update for each client \( k \) using the aggregated weights is then computed as:
\begin{equation}
W_k = M_k \odot W_{\text{global}}^{q+1}
\end{equation}

This process repeats for \( Q \) rounds until convergence is achieved.
\end{itemize}

\subsection{Model Complexity Analysis}

\begin{table*}[ht]
\centering
\scriptsize
\renewcommand{\arraystretch}{0.88}
\caption{Overview of selected IoT datasets }
\label{tab:datasets}
\begin{tabular}{p{0.25cm}p{1cm}p{8.8cm}p{5.9cm}}
\toprule
\textbf{Year} & \textbf{Dataset}  & \textbf{Attack Categories} & \textbf{Notable Characteristics} \\
\midrule

2020 & Ton\_IoT  &  Backdoor, DDoS, DoS, Injection Attacks, Man-in-the-middle (MITM) attacks,  Normal, Password Cracking, Ransomware,  Scanning, and cross-site scripting (XSS). & A recent IoT-specific dataset built on real-world interactions; particularly suitable for deploying deep transfer learning (DTL) models. \\
\midrule
2022 & X-IIoTID  & Brute Force, Command and Control (C\&C), Dictionary,  Discovering Resources,   Exfiltration, Fake Notification,  False Data Injection,  Generic Scanning, MQTT Cloud Broker Subscription, MITM Attacks, Modbus Register reading, Normal, Ransom Denial of Service (RDoS),  Reverse Shells, Scanning  Vulnerability,  TCP Relay,   Crypto-Ransomware,  Fuzzing,  Insider Malicious. & A comprehensive and up-to-date dataset reflecting modern IIoT environments, protocols, device behaviors, and sophisticated attack scenarios. \\
\midrule
2024  & IDSIoT2024 & ARP Poisoning, Backdoor,  ICMP Flood,  ICMP Redirect,  Normal, Password Cracking, Port Scanning, SQL Injection,  SYN Flood,  Smurf, UDP Flood, Vulnerability Scan. & A recent, realistic IoT dataset covering diverse attack types. Ideal for training and evaluating ML/DL-based IDS models. \\
\bottomrule
\end{tabular}%
\end{table*}

The objective of the model complexity analysis is to evaluate the computational and structural demands of our proposed approach. This step is crucial, as reducing model complexity directly impacts energy consumption, inference speed, and the feasibility of deployment on IoT devices. CNN models were selected for their proven effectiveness in extracting spatial features, which are essential for the target tasks. A detailed analysis of the proposed CNN model, both before and after pruning, is provided.

\textbf{Model Parameter Calculation:} The total number of parameters in the CNN before pruning is mainly composed of two parts: convolutional (CONV) layers and fully connected (FC) layers. Although each layer includes bias terms, their contribution is negligible compared to the number of weight parameters.  
\\[0.1cm]
\textit{CONV Layers:} In a multi-layer 1D CNN, the total parameters are dominated by weights. For each CONV layer \( i \), the kernel size \( K_i \), the number of input channels \( C_{\text{in},i} \), and the number of output channels \( C_{\text{out},i} \) determine the parameter count. The total number of weights in all CONV layers is:
\begin{equation}
NP_{C,\text{unp}} = \sum_{i=1}^{L_C} \left( K_i \times C_{\text{in},i} \times C_{\text{out},i} \right)
\end{equation}

where \( L_C \) represents the total number of CONV layers.\\[0.1cm] 
\textit{FC Layers:} For FC layers, if the \( j \)-th layer has \( N_{\text{in},j} \) input neurons and \( N_{\text{out},j} \) output neurons, the total weight parameters in all FC layers is:
\begin{equation}
NP_{FC,\text{unp}} = \sum_{j=1}^{L_F} \left( N_{\text{in},j} \times N_{\text{out},j} \right)
\end{equation}

where \( L_F \) denotes the total number of FC layers.

Given that the number of bias terms is significantly smaller than the number of weight parameters, it can be considered negligible. Thus, the total number of parameters in the CNN Model before pruning is approximated as:
\begin{equation}
NP_{\text{unp}} = \sum_{i=1}^{L_C} \left( K_i \times C_{\text{in},i} \times C_{\text{out},i} \right) + \sum_{j=1}^{L_F} \left( N_{\text{in},j} \times N_{\text{out},j} \right)
\end{equation}

After weight pruning, the number of parameters in in the CNN Model  is given by:
\begin{equation}
NP_{\text{p}} = (1 - \rho  )  \times NP_{\text{unp}}
\end{equation}
where \( \rho \) represents the pruning ratio.

\textbf{Model Computational Cost Calculation:} In DL networks, computational complexity is often measured using one key metric: FLOPs (Floating Point Operations) \cite{10830510}. FLOPs refer to the total number of floating-point operations (both multiplications and additions) required for model inference.\\[0.1cm]
\textit{CONV Layers:} The total number of operations in a CONV layer includes both multiplications and additions. Here, \( L_{\text{out},i} \) denotes the length of the output feature map for layer \( i \). The FLOPs for the CONV layers can be calculated as:
\begin{equation}
FLOPs_{c,\text{unp}} = \sum_{i=1}^{L_C} \left( 2 \times K_i \times C_{\text{in},i} \times L_{\text{out},i} \times C_{\text{out},i} \right)
\end{equation}

In DL networks, a "Multi-Add" is often considered as one floating-point operation (counting both multiplication and addition together). In this case, the total number of operations could be halved.\\[0.1cm]
\textit{FC Layers:} The computational complexity of a FC layer depends on its number of input and output neurons. The total number of operations required by the FC layer is:
\begin{equation}
FLOPs_{FC,\text{unp}} = \sum_{j=1}^{L_F} \left( 2 \times N_{\text{in},j} \times N_{\text{out},j} \right)
\end{equation}

Thus, the total computational cost of the CNN model, in terms of FLOPs before pruning, is the sum of the operations from both the CONV and FC layers:
\begin{equation}
FLOPs_{\text{unp}} = FLOPs_{C,\text{unp}} +  FLOPs_{FC,\text{unp}}
\end{equation}

After pruning, the number of FLOPs in the CNN model is adjusted based on \( \rho \). Since pruning reduces the number of parameters, it also affects the total number FLOPs. The number of FLOPs after pruning for the CONV layers is given by:
\begin{equation}
FLOPs_{\text{p}} = (1 - \rho)  \times  FLOPs_{\text{unp}} 
\end{equation}

\subsection{Problem Formulation}
 We introduce an optimization approach designed to maximize the model’s performance across all clients, while taking into account both maximizing accuracy and minimizing energy consumption during inference. The approach places particular emphasis on pruning strategies. For \(N\) clients, the objective can be formulated as:
\begin{equation}
\max \left\{
    \frac{ \alpha_1 \, }{N} \sum_{i=1}^N\mathrm{Acc}_i 
    + \frac{\alpha_2}{\frac{1}{N} \sum_{i=1}^N \mathrm{E}_{\mathrm{con},i}} 
\right\}
\end{equation}

The optimization is achieved through an objective function that incorporates the following two factors:
\begin{itemize}[left=0pt]
   \item Accuracy: For client \(i\), this factor is represented by the sum of correct predictions for its local dataset \(D_i\), weighted by the coefficient \(\alpha_1\). The goal is to maximize the model's precision while controlling the pruning process.
\begin{equation}
\text{Acc}_i = \frac{1}{|D_i|} \sum_{j \in D_i} \mathbf{1}(\hat{y}_j = y_j)
\end{equation}

\item Energy Consumption: The energy consumption of a model during inference is a critical metric for assessing its efficiency, especially in resource-constrained environments. As highlighted in \cite{10398587}, total energy consumption is influenced by two main factors: the energy required for performing computations and the energy required for memory access.

The computational energy cost is primarily determined by the number of FLOPs, while the memory access energy cost depends on the model size. The model size can be estimated using the following formula:
\begin{equation}
\text{Model Size} = \text{NP} \times B
\end{equation}

where \( \text{NP} \) is the number of parameters and \( B \) is the number of bytes used per parameter.

The overall energy consumption for client \( i \) is given by:
\begin{equation}
E_{\text{con}, i} = (\text{FLOPs}_i \times E_{\text{FLOP}}) + (\text{Model Size}_i \times E_{\text{access}})
\end{equation}

where \( E_{\text{FLOP}} \) and \( E_{\text{access}} \) denote the energy cost per arithmetic operation and per unit of memory access, respectively. 
\end{itemize}
Before pruning, we can rewrite our previous optimization function as:
\begin{equation}
\max \left\{ 
    \frac{\alpha_1 \,}{N} \sum_{i=1}^{N}  \mathrm{Acc}_i^{\mathrm{unp}} 
    + \frac{\alpha_2}{\frac{1}{N} \sum_{i=1}^{N} E_{\mathrm{con},i}^{\mathrm{unp}}}
\right\}
\end{equation}

By removing certain weights during pruning, the complexity decreases. This reduction in complexity leads to lower energy consumption, which can be expressed after pruning for client \(i\) through the pruning term \(\rho_i\) as follows:
\begin{equation}
\max \left\{ 
    \frac{\alpha_1 \,}{N} \sum_{i=1}^{N}  \mathrm{Acc}_i^p 
    + \frac{\alpha_2}{\frac{1}{N} \sum_{i=1}^{N} E_{\mathrm{con},i}^{p}} 
\right\}
\end{equation}

Where \(\rho_i\) is the pruning percentage, representing the fraction of parameters removed from the local model. The energy consumption after pruning is given by:
\[
E_{\text{con},i}^p = (1 - \rho_i)  E_{\text{con},i}^{\text{unp}}
\]

An approximate relationship between  \( \rho \) and the accuracy loss after pruning can be expressed as follows \cite{10.5555/2969239.2969366}:
\begin{figure}[t!]
  \centering    \includegraphics[width=1\columnwidth]{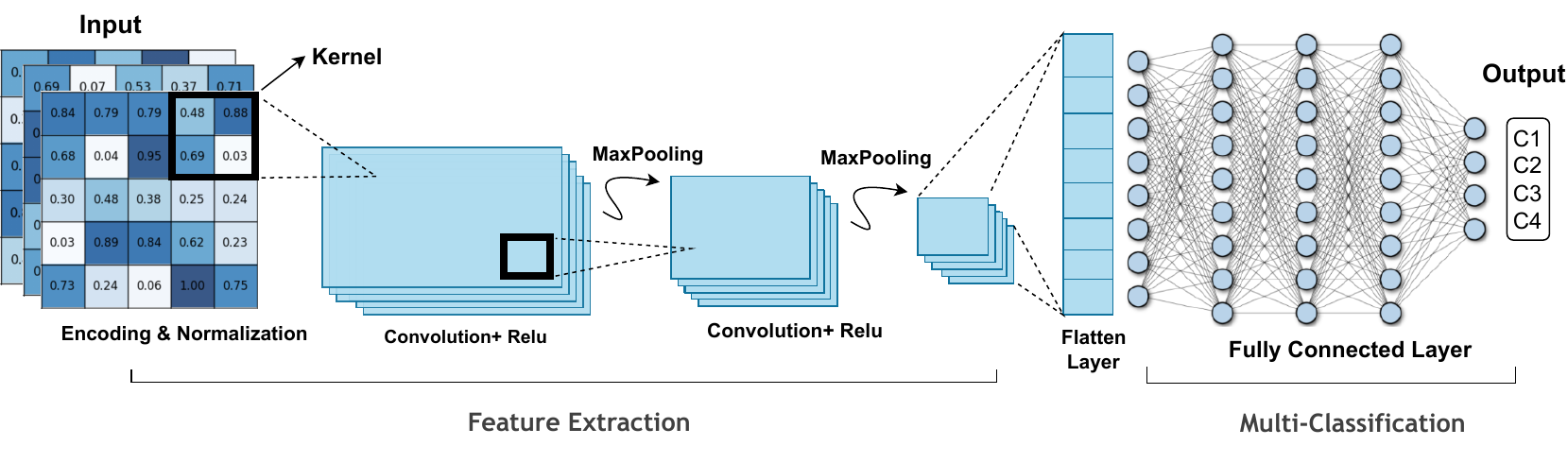}
    \caption{Proposed model architecture for IDS}
    \label{fig:Model Architecture1} 
\end{figure}

\begin{equation}
\text{Acc}_i^p \approx \text{Acc}_i^{\text{unp}} \left(1 - \beta_i \cdot e^{\lambda_i \rho_i} \right)
\end{equation}
where \( \beta_i > 0 \) is the pruning impact factor specific to client \( i \), and \( \lambda_i > 0 \) is the exponential decay rate for client \( i \); both parameters are related to the structure and behavior of the CNN model.

The multi-objective optimization becomes:
\begin{equation}
\begin{aligned}
\max_{\rho} \Bigg\{ 
& \frac{\alpha_1}{N} \sum_{i=1}^N  \mathrm{Acc}_i^{\mathrm{unp}} 
     \left(1 - \beta_i e^{\lambda_i \rho_i} \right) \\
& + \frac{\alpha_2}{\frac{1}{N} \sum_{i=1}^N (1 - \rho_i) E_{\mathrm{con},i}^{\mathrm{unp}}}
\Bigg\}
\label{eq:objectif_normalise}
\end{aligned}
\end{equation}

Subject to:
\begin{equation}
0 \leq \rho_i \leq 1,
\end{equation}
\begin{equation}
\text{Acc}_i^p \geq \text{Acc}_i^{\text{unp}} - \delta, \quad \forall i,
\end{equation}
where \( \delta \) is a small positive value representing the maximum allowable accuracy degradation.

\subsection{Using Deep Reinforcement Learning For Optimization}
\begin{figure*}[!t]
    \centering
    \begin{subfigure}[t]{0.22\textwidth}
        \centering
        \includegraphics[width=\linewidth]{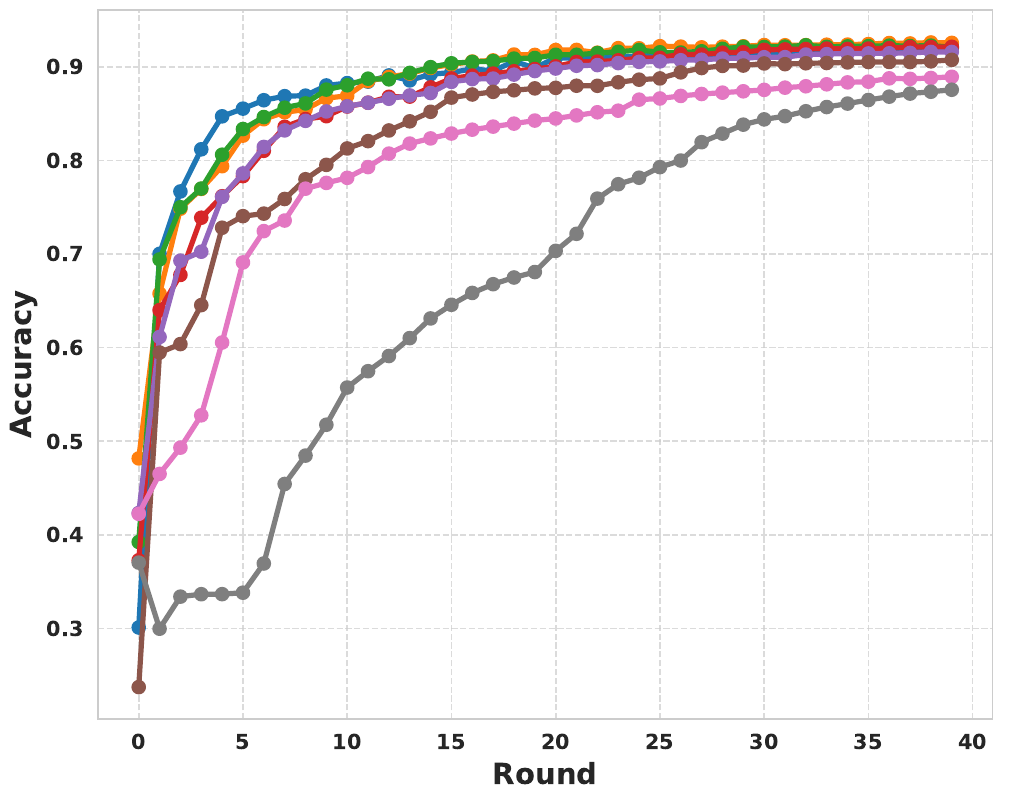}
        \caption{  Ton\_IoT: Accuracy vs. Rounds }
        \label{fig:img1}
    \end{subfigure}
    \hfill
    \begin{subfigure}[t]{0.22\textwidth}
        \centering
        \includegraphics[width=\linewidth]{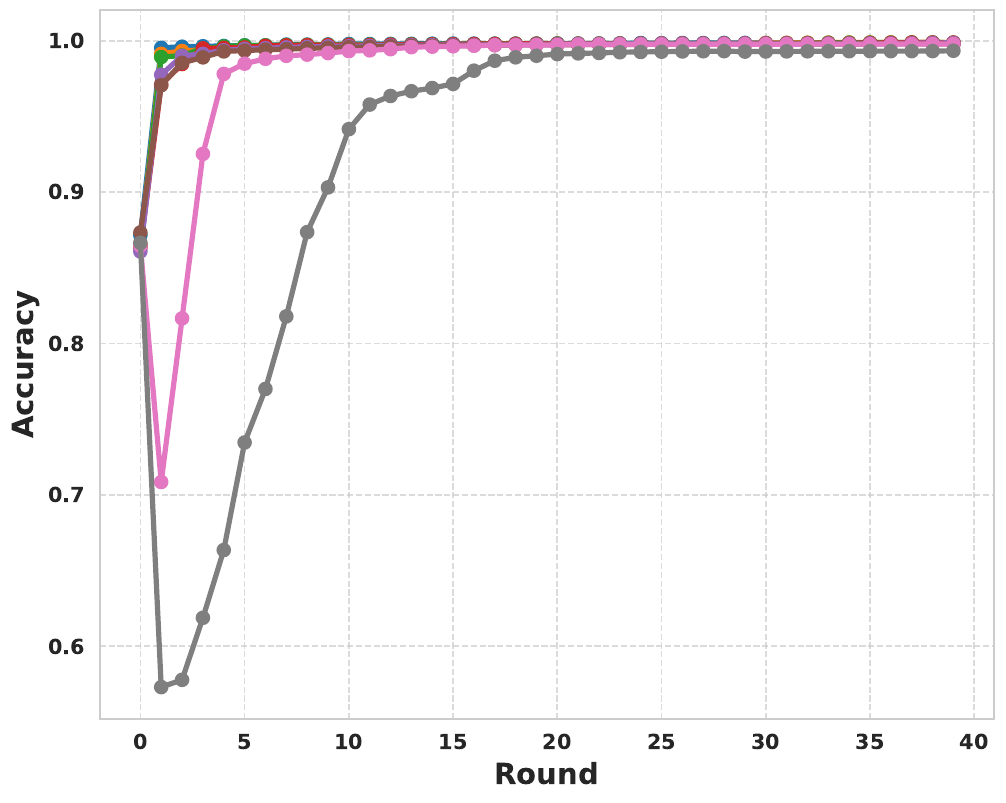}
        \caption{IDSIoT2024: Accuracy vs. Rounds }
        \label{fig:img2}
    \end{subfigure}
    \hfill
    \begin{subfigure}[t]{0.28\textwidth}
        \centering
        \includegraphics[width=\linewidth]{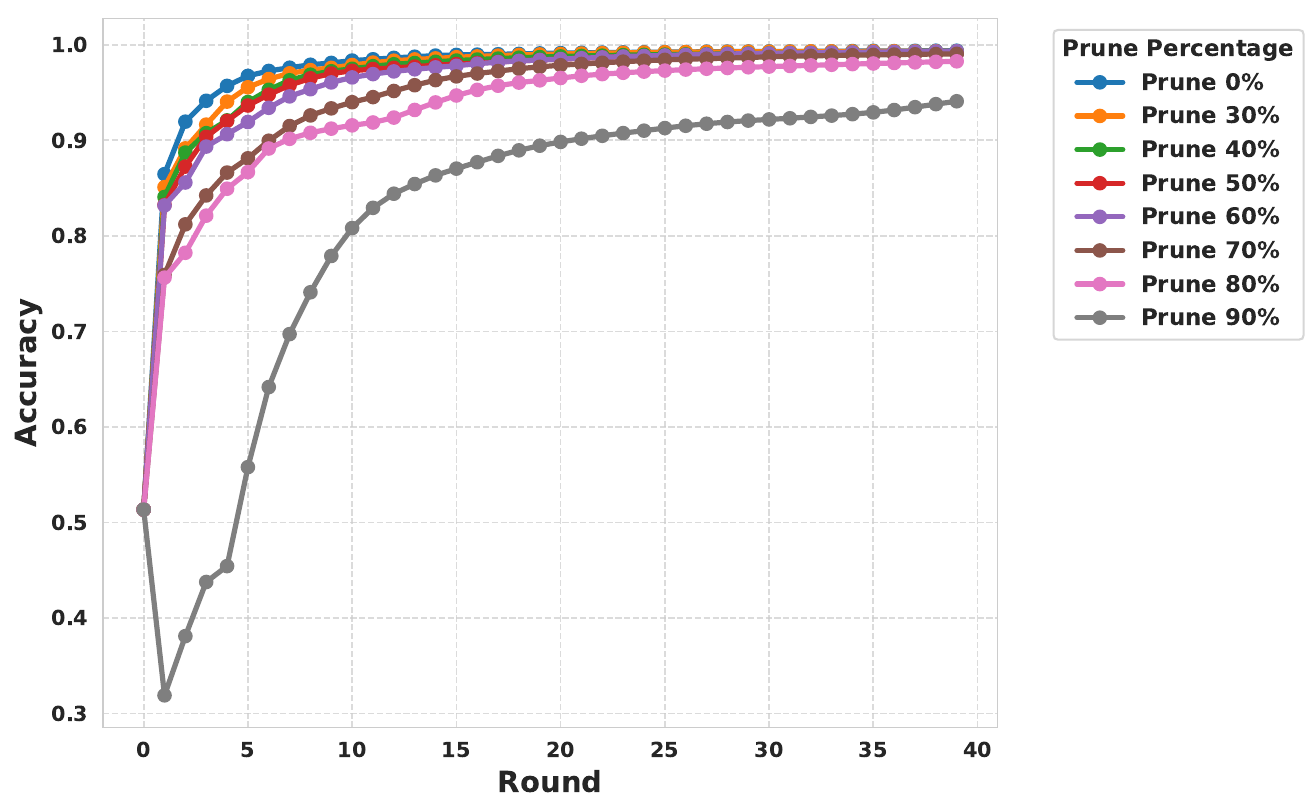}
        \caption{X-IIoT-ID: Accuracy vs. Rounds }
        \label{fig:img3}
    \end{subfigure}
    \caption{Impact of pruning on model accuracy across three IoT datasets under IID settings using FedAvg}

    \label{fig:ACC all_images}
\end{figure*}

\begin{figure*}[!t]
    \centering
    \begin{subfigure}[t]{0.23\textwidth}
        \centering
        \includegraphics[width=\linewidth]{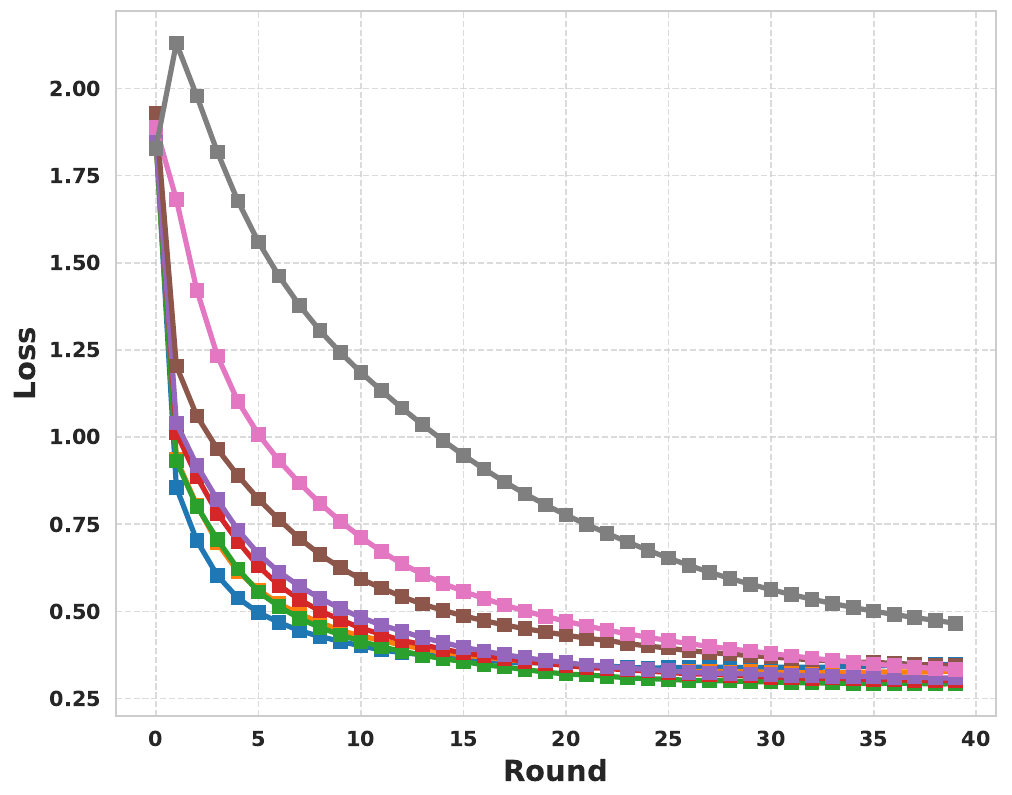}
        \caption{ Ton\_IoT: Loss vs. Rounds}
        \label{fig:loss1}
    \end{subfigure}
    \hfill
    \begin{subfigure}[t]{0.23\textwidth}
        \centering
        \includegraphics[width=\linewidth]{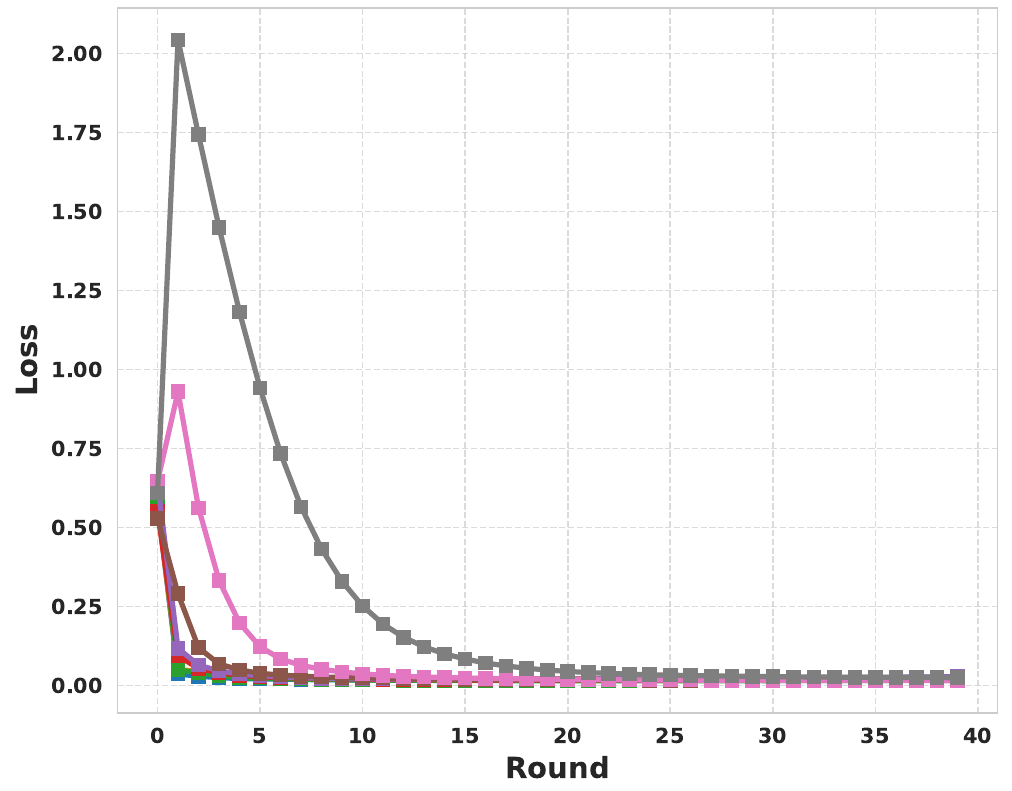}
        \caption{IDSIoT2024: Loss vs. Rounds}
        \label{fig:loss2}
    \end{subfigure}
    \hfill
    \begin{subfigure}[t]{0.30\textwidth}
        \centering
        \includegraphics[width=\linewidth]{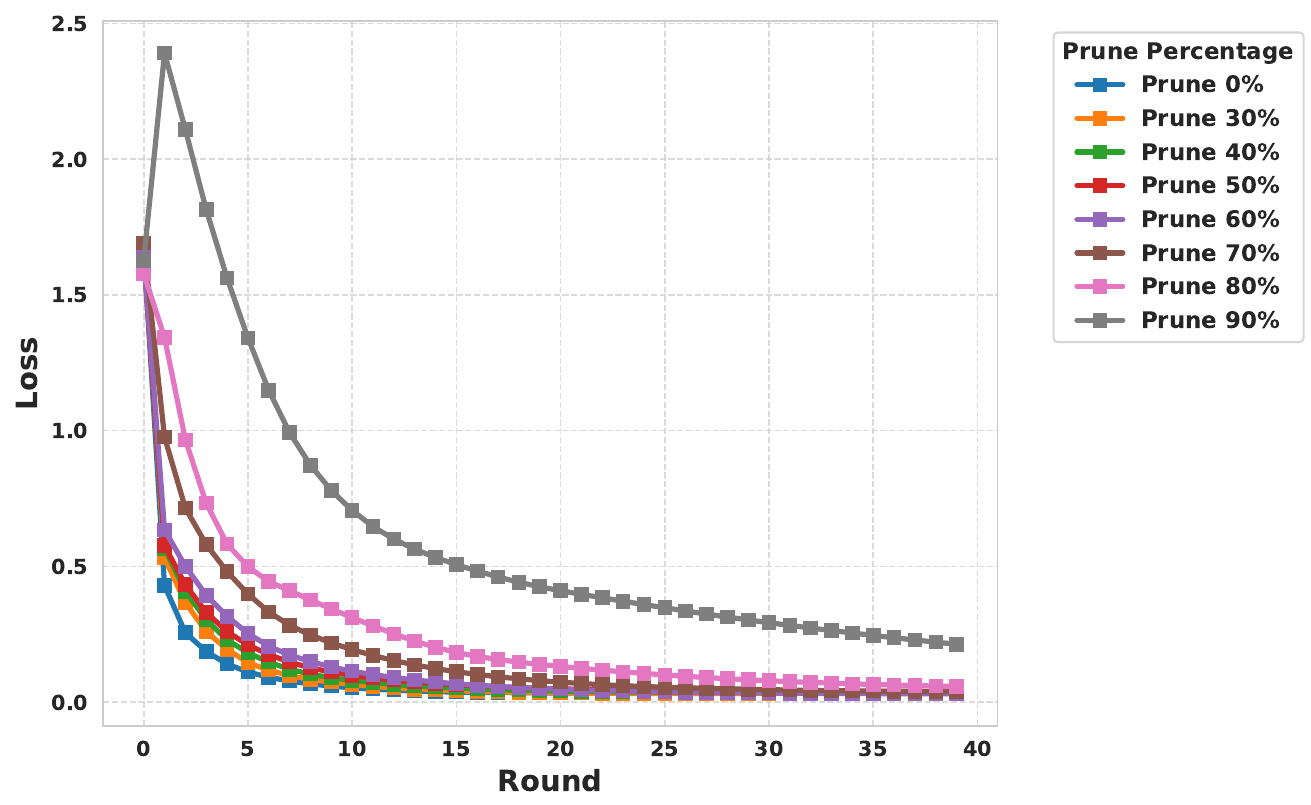}
        \caption{X-IIoT-ID: Loss vs. Rounds}
        \label{fig:loss3}
    \end{subfigure}
    \caption{Impact of pruning on model loss across three IoT datasets under IID settings using FedAvg}

    \label{fig:loss_all_images}
\end{figure*}

In this work, we used DRL to solve the optimization problem that we described earlier. Our goal was to find the best trade-off between model accuracy and energy consumption when pruning neural networks in a FL setup. Instead of using fixed rules or traditional optimization methods \cite{nocedal2006numerical}, we let a DRL agent learn how to choose the appropriate pruning ratios. The agent interacts with the system and receives feedback on how well the model performs after pruning. Based on this feedback, the agent improves its strategy over time. Therefore, the agent learns which pruning decisions give the best trade-off between high accuracy and low energy cost.

To implement this, we used the \textit{Stable Baselines3} framework~\cite{stable-baselines}, which is a well-known open source library for DRL. We selected the PPO algorithm~\cite{articlePPO} since it works well with continuous actions and is stable during training. PPO can help the agent learn better pruning decisions for multiple nodes and communication rounds.
Importantly, we resolved this optimization problem offline, allowing the agent to train in a simulated environment before deployment. This offline training enables robust learning without interfering with real-time FL operations. By integrating DRL, our system becomes more adaptable, as it tailors the pruning strategy for each client based on their data distribution and resource constraints. This results in a more efficient and scalable IDS, especially suited for real-world IoT environments where data heterogeneity and energy limitations are common.

\section{EXPERIMENTAL SETUP AND PERFORMANCE EVALUATION} \label{C}
This section presents the essential details of the experiments carried out, including the datasets used, experimental configurations, and performance assessment.

\subsection{Selected Datasets}
We evaluate the performance of the \textit{OptiFLIDS} framework using three diverse and widely recognized datasets: Ton\_IoT \cite{9189760}, X-IIoTID \cite{9504604}, and IDSIoT2024 \cite{10914935}. These datasets are particularly suitable for assessing NIDS, as they include recent IoT-related attacks and realistic external communication traffic. Table~\ref{tab:datasets} summarizes the attacks present in each dataset.

\subsection{Experimental Settings}

The experiments were conducted on a single node within the African Supercomputing Center (ASCC) HPC cluster, which is equipped with four NVIDIA A100 SXM4 GPUs, each having 80 GB of memory. Our complete code repository available on at \url{https://github.com/SAIDAELOUARDI23/OptiFLIDS-.git}.
To start with, assuming a secure channel is established between the NIDS global server and the clients, we define the model architecture and initialize its parameters. The proposed model is designed for the detection of intrusions in IoT networks, using CONV layers to extract relevant features from the network flow data, followed by FC  layers for classification. As illustrated in Fig.~\ref{fig:Model Architecture1}, the architecture comprises two main components: a feature extractor, which captures hierarchical patterns from raw IoT data using CONV layers, and a classifier, which processes these features to categorize the input into predefined attack or benign classes.

\begin{itemize}[left=0pt]
  \item \textit{Weight Initialization:} The CONV layers utilize Kaiming He initialization to address vanishing or exploding gradients. The FC layers are initialized with weights drawn from a normal distribution (mean = 0, standard deviation = 0.01).

\item \textit{Configurations:} The OptiFLIDS  framework was implemented using Python 3.9.21, along with several key libraries, including PyTorch, NumPy, Pandas, Matplotlib, and Scikit-learn.  For each dataset ( Ton\_IoT, X-IIoTID, and IDSIoT2024), 80\% of the data was used for training, while 20\% was reserved for final evaluation. The training process involved local models that were trained over 20 epochs with a learning rate of 0.001. To address client drift, the FedProx algorithm was applied with a proximal term coefficient Mu set to 0.001.

\item \textit{Communication rounds:} During the FL phase, the number of communication rounds \(Q\) between the server and clients was fixed at 40, as no significant improvements were observed beyond this point. The proposed framework was evaluated with 10, 60, and 100 clients for the Ton\_IoT, X-IIoTID, and IDSIoT2024 datasets, respectively. For both IID and non-IID data distribution scenarios, client datasets were generated using a Normalized Gamma Distribution, with \(\alpha = 10\) for non-IID data and \(\alpha = 1000000\) for IID data across all datasets.

\end{itemize}
In the model pruning process, we tested different sparsity levels, ranging from 10\% to 90\%. We assessed performance using several metrics, including accuracy, loss, and FLOPs.


\begin{figure*}[!th]
    \centering
    \begin{subfigure}[t]{0.23\textwidth}
        \centering
        \includegraphics[width=\linewidth]{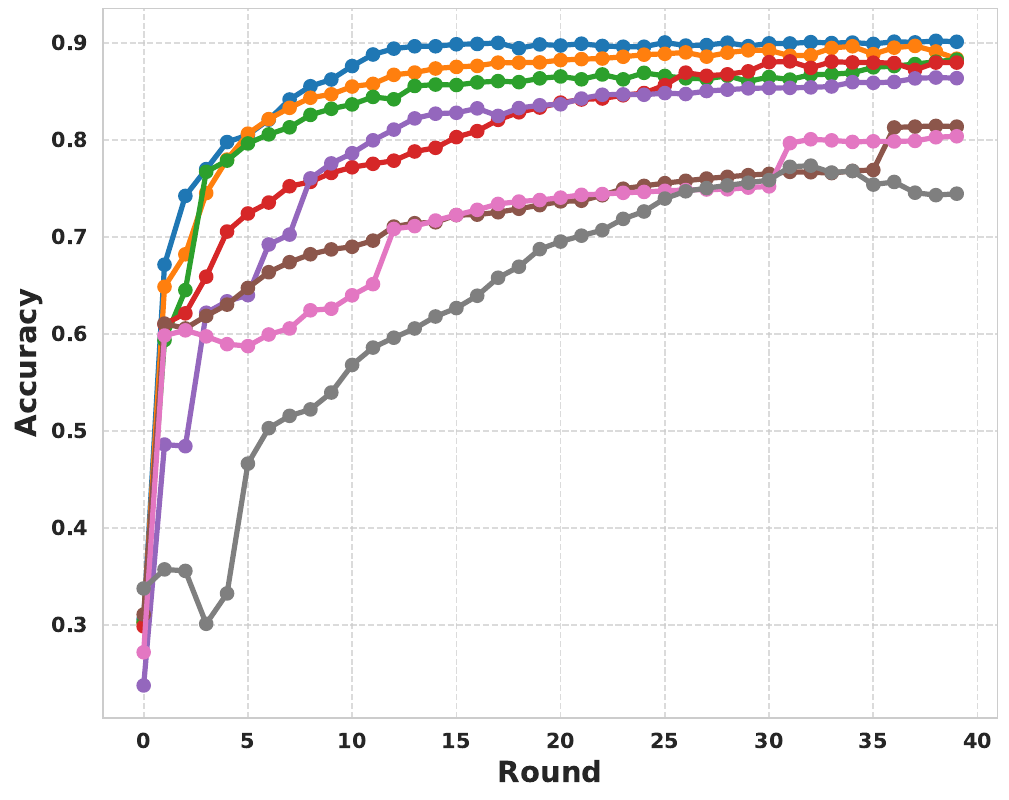}
        
        \caption{ Ton\_IoT: Accuracy vs. Rounds}
        \label{fig:noniid_1}
    \end{subfigure}
    \hfill
    \begin{subfigure}[t]{0.23\textwidth}
        \centering
        \includegraphics[width=\linewidth]{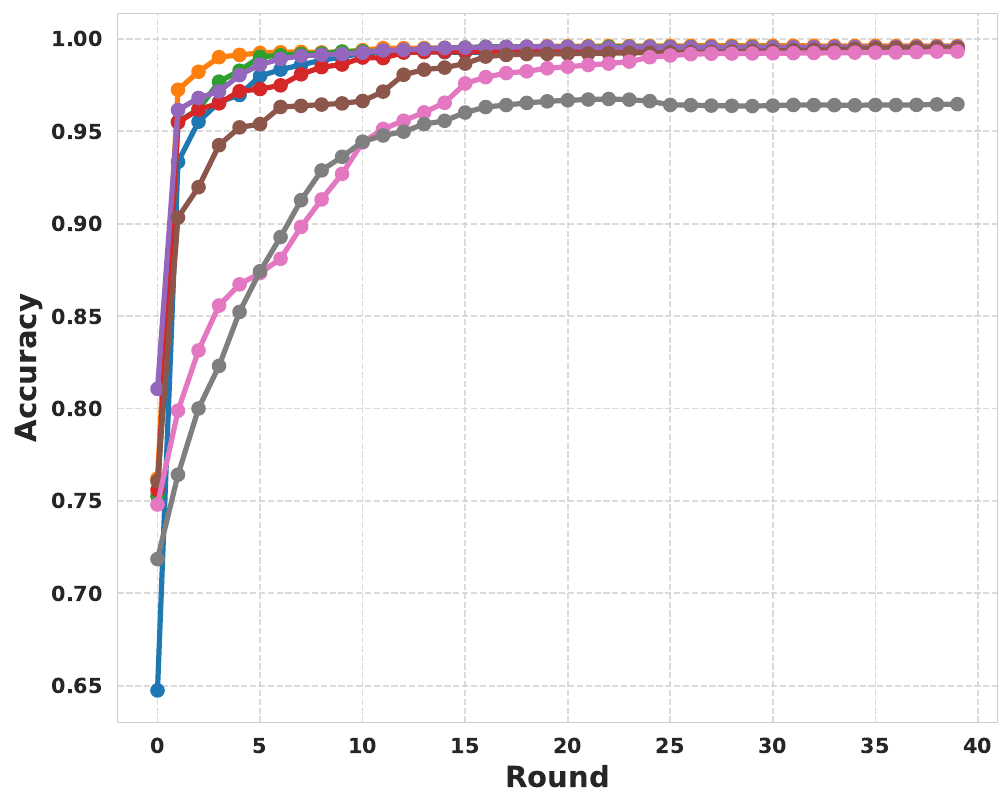}
        \caption{IDSIoT2024: Accuracy vs. Rounds}
        \label{fig:noniid_2}
    \end{subfigure}
    \hfill
    \begin{subfigure}[t]{0.29\textwidth}
        \centering
        \includegraphics[width=\linewidth]{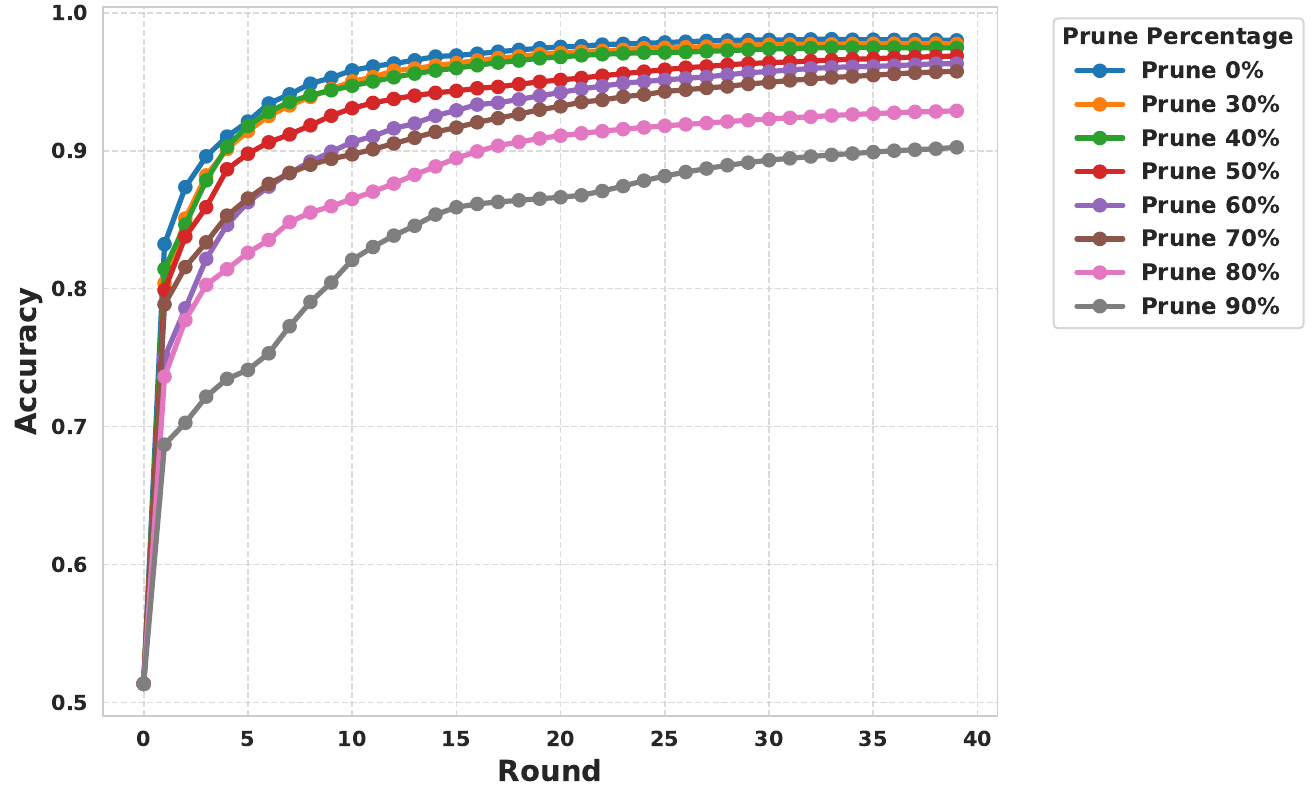}
        \caption{X-IIoTID: Accuracy vs. Rounds}
        \label{fig:noniid_3}
    \end{subfigure}
    \caption{Impact of pruning on model accuracy across three IoT datasets under Non-IID settings using FedAvg}

    \label{fig:noniid_ACC_all_images}
\end{figure*}

\begin{figure*}[!th]
    \centering
    \begin{subfigure}[t]{0.24\textwidth}
        \centering
        \includegraphics[width=\linewidth]{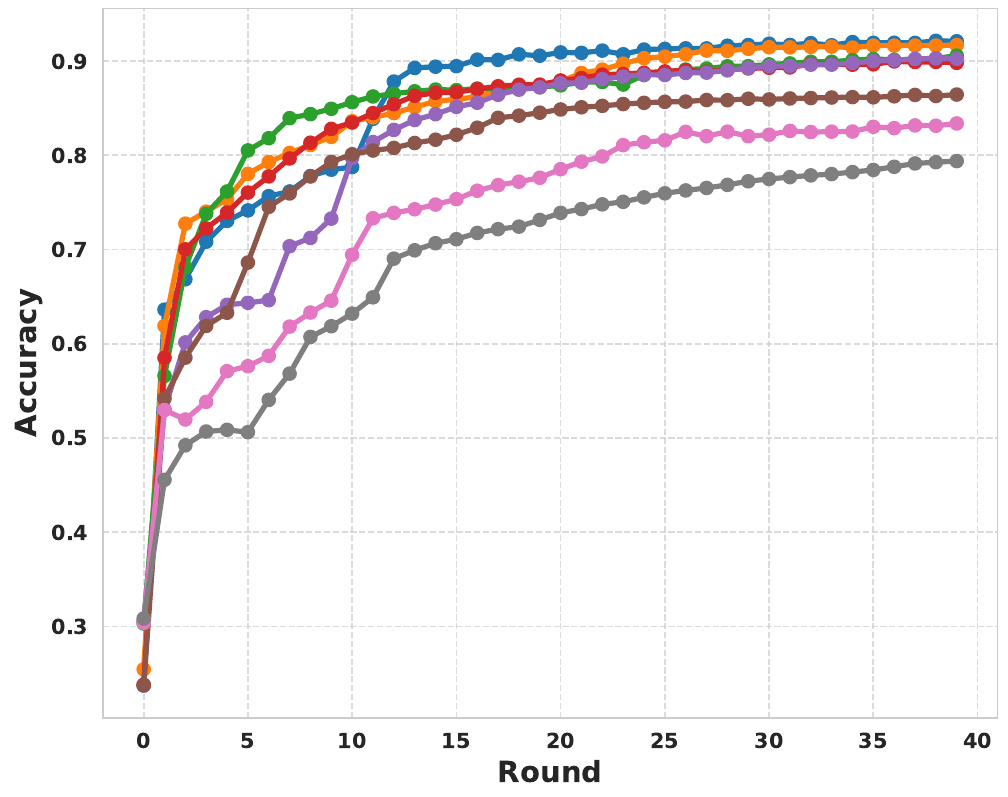}
        \caption{ Ton\_IoT: Accuracy vs. Rounds}
        \label{fig:noniid_fedprox_img1}
    \end{subfigure}
    \hfill
    \begin{subfigure}[t]{0.24\textwidth}
        \centering
        \includegraphics[width=\linewidth]{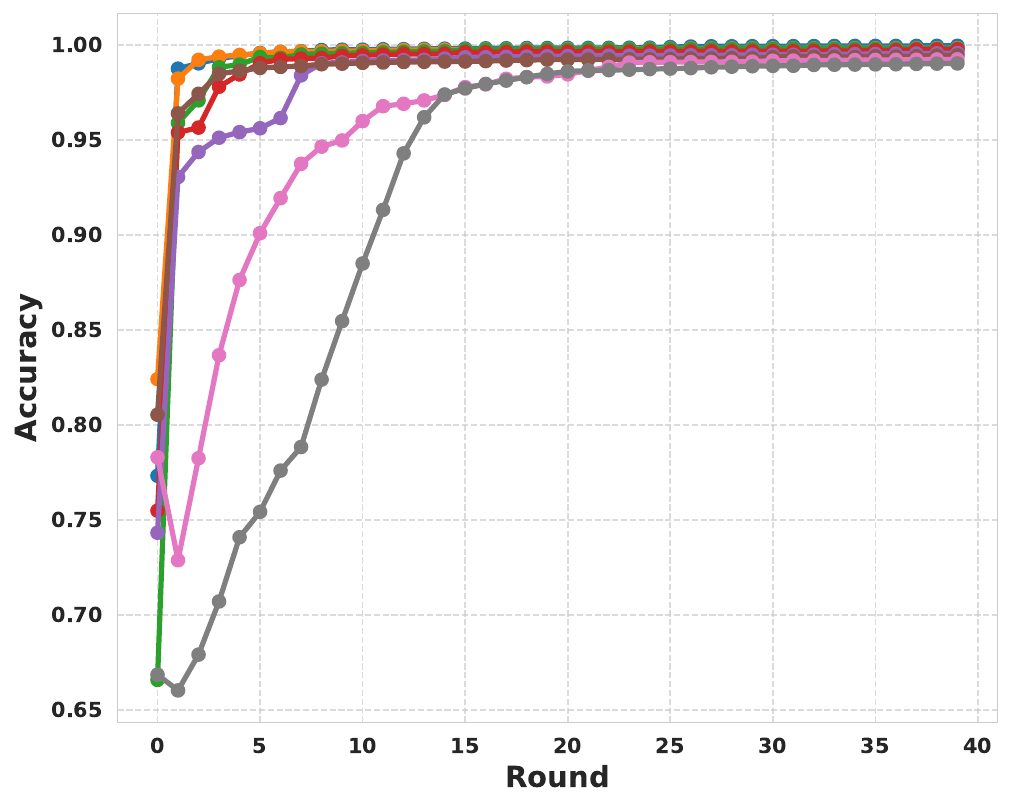}
        \caption{IDSIoT2024: Accuracy vs. Rounds}
        \label{fig:noniid_fedprox_img2}
    \end{subfigure}
    \hfill
    \begin{subfigure}[t]{0.31\textwidth}
        \centering
        \includegraphics[width=\linewidth]{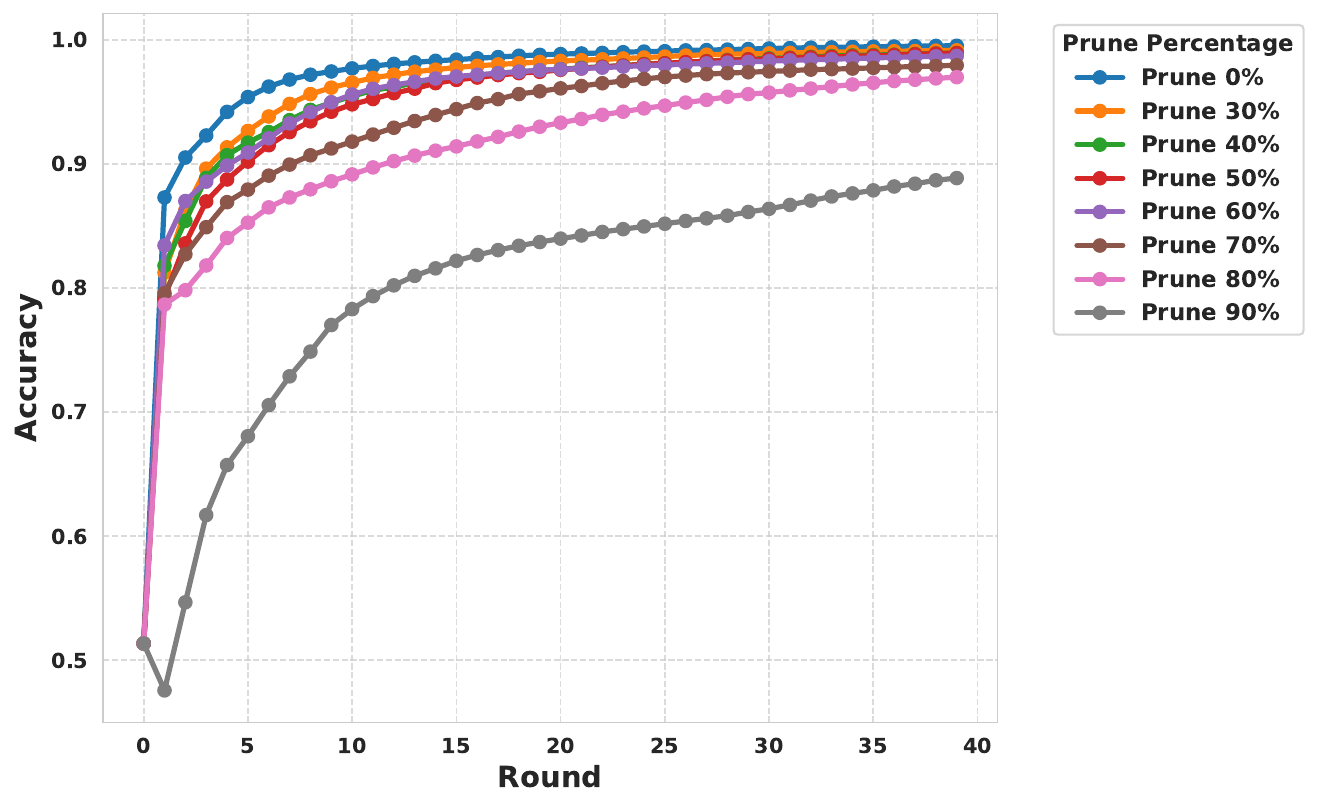}
        \caption{X-IIoTID: Accuracy vs. Rounds}
        \label{fig:noniid_fedprox_img3}
    \end{subfigure}
    \caption{Impact of pruning on model accuracy across three IoT datasets under Non-IID settings using FedProx}
    \label{fig:noniid_ACC_all_images_fedprox}
\end{figure*}

\subsection{Results Evaluation And Discussion}

\begin{figure*}[!t]
    \centering
    \begin{subfigure}[t]{0.23\textwidth}
        \centering
        \includegraphics[width=\linewidth]{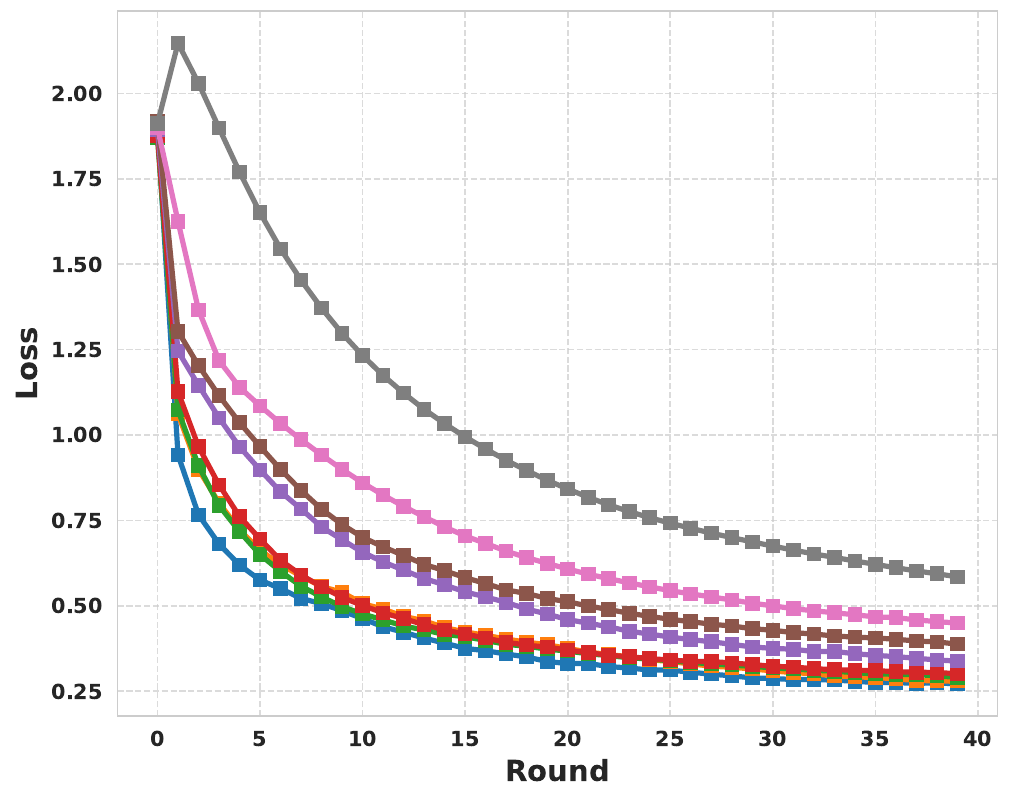}
        \caption{ Ton\_IoT: Loss vs. Rounds}
        \label{fig:noniid_loss_fedprox_img1}
    \end{subfigure}
    \hfill
    \begin{subfigure}[t]{0.23\textwidth}
        \centering
        \includegraphics[width=\linewidth]{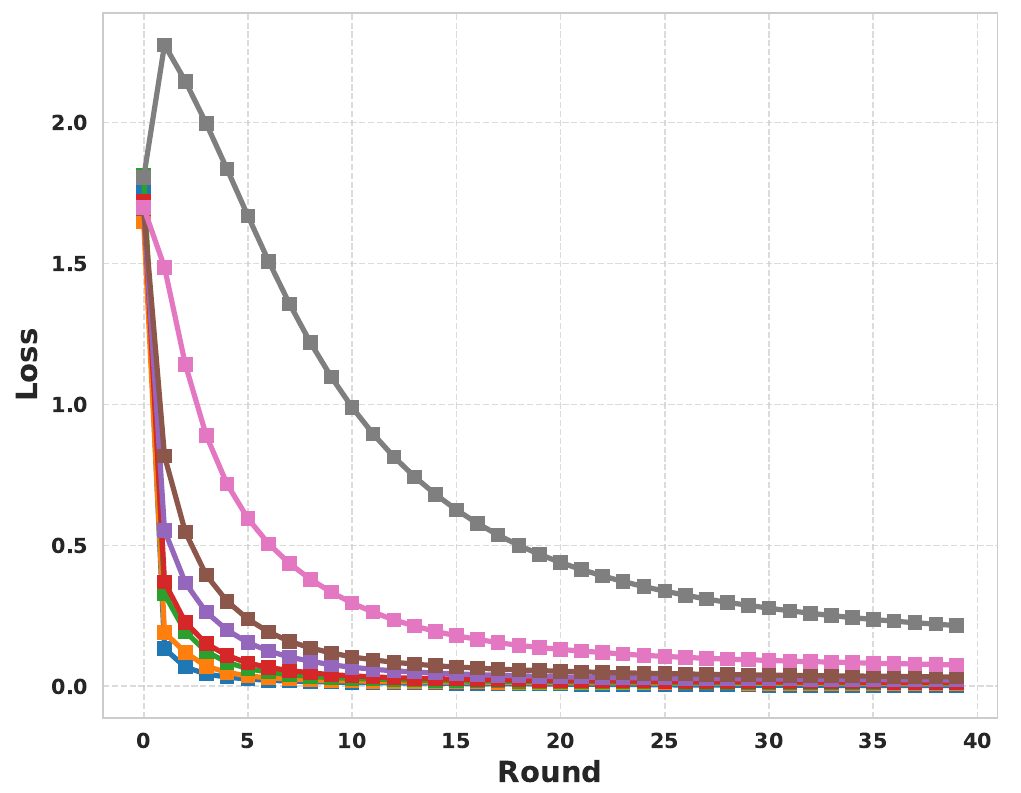}
        \caption{IDSIoT2024: Loss vs. Rounds}
        \label{fig:noniid_loss_fedprox_img2}
    \end{subfigure}
    \hfill
    \begin{subfigure}[t]{0.30\textwidth}
        \centering
        \includegraphics[width=\linewidth]{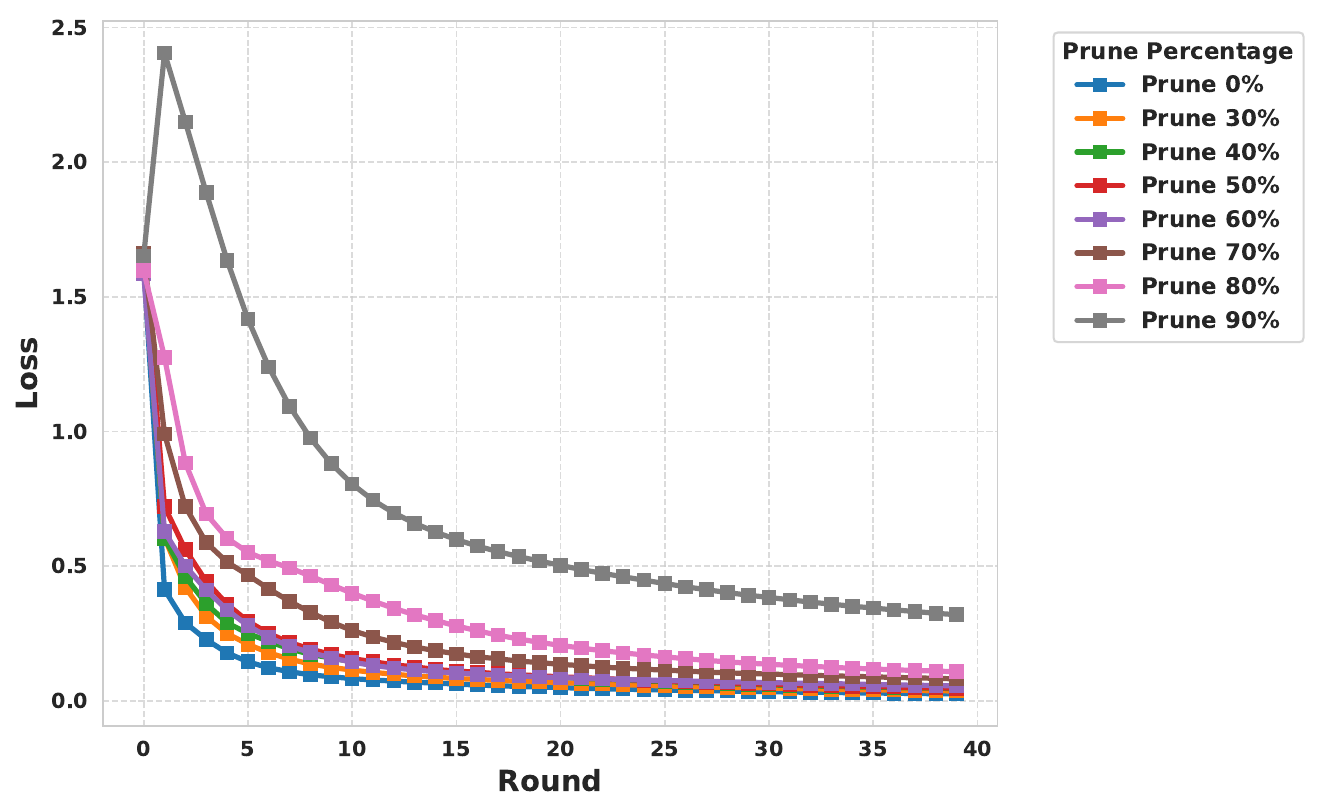}
        \caption{X-IIoTID: Loss vs. Rounds}
        \label{fig:noniid_loss_fedprox_img3}
    \end{subfigure}
    \caption{Impact of pruning on model loss across three IoT datasets under non-IID settings using FedProx}
    \label{fig:noniid_loss_fedprox_all_images}
\end{figure*}

\begin{figure*}[ht]
    \centering
    \begin{subfigure}[t]{0.29\textwidth}
        \centering
        \includegraphics[width=\linewidth]{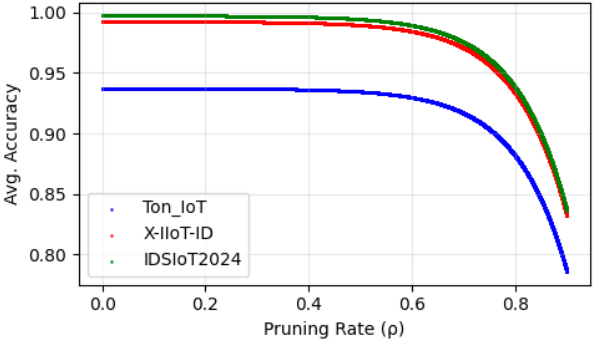}
        \label{fig:}
    \end{subfigure}
    \hfill
    \begin{subfigure}[t]{0.31\textwidth}
        \centering
        \includegraphics[width=\linewidth]{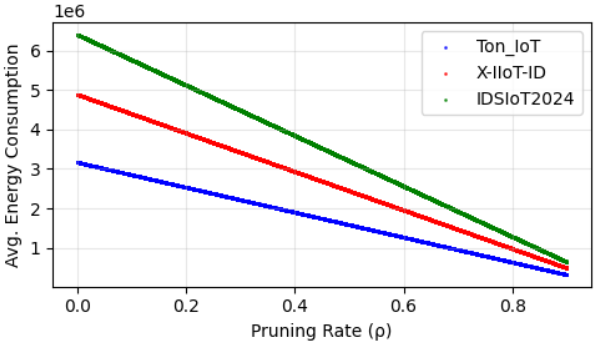}
        \label{fig:2}
    \end{subfigure}
    \hfill
    \begin{subfigure}[t]{0.29\textwidth}
        \centering
        \includegraphics[width=\linewidth]{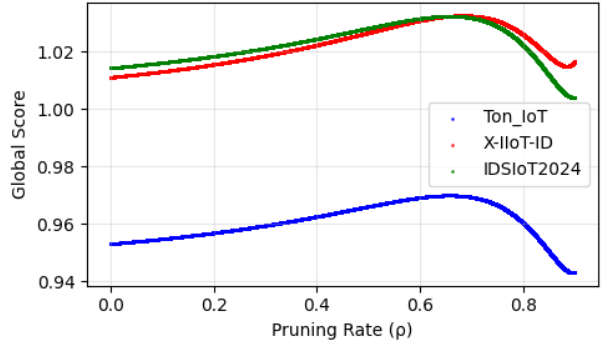}
        \label{fig:no3}
    \end{subfigure}
    \caption{Optimization results using DRL: Impact of Pruning on accuracy, energy consumption, and score}
    \label{fig:3}
\end{figure*}

\begin{table*}[!t]
\caption{Performance and Efficiency Metrics for Different Datasets}
\centering
\begin{tabular}{|c|c|c|c|c|c|c|c|}
\hline
\textbf{Dataset} & \textbf{Nb of Parameters} & \textbf{FLOPs} & \textbf{Energy(pJ)} & \textbf{Unpruned Accuracy Range} & \textbf{Best Rho} & \textbf{Best Score} & \textbf{Nb of Clients} \\
\hline
\textbf{Ton\_IoT}   & 190218 & 1378560& 3171152.39 & 93--94\% & 65.75\% &0.9699 & 10 \\
\hline
\textbf{X-IIoT-ID} & 289682 & 2126976 & 4892751.232 & 98--99.40\% &68.36\% & 1.0325 & 60 \\
\hline    
\textbf{IDSIoT2024} & 370698 & 2788224 & 6413819.392 & 96--99.90\% & 65.85\% &1.0323 & 100 \\           
\hline
\end{tabular}
\label{tab:DRL}
\end{table*}

Fig.\ref{fig:ACC all_images} shows how pruning affects model accuracy during training on three different IoT datasets:  Ton\_IoT, IDSIoT2024, and X-IIoTID. The training is done under IID settings using the FedAvg algorithm. We compare how the accuracy changes over 40 communication rounds for different pruning levels, from no pruning (0\%) up to 90\%.

For the  Ton\_IoT (Fig.~\ref{fig:img1}) and X-IIoTID (Fig.~\ref{fig:img3}) datasets, the models demonstrate similar behavior when exposed to pruning. In both cases, accuracy remains close to the original model up to moderate pruning levels (30\%–60\%). The training is stable, and the models gradually improve over the rounds, indicating a certain weights are not important and allows for effective pruning. However, Beyond 60\%, the accuracy decreases significantly. Especially at pruning levels of 90\%, the learning process becomes less efficient, and the models tend to converge poorly. This suggests that pruning beyond this point removes too much useful information, making it harder for the models to learn.

In comparison to other datasets, the IDSIoT2024 dataset (Fig. \ref{fig:img2}) appears to enhance the model's resilience to pruning. The dataset may contain duplicate samples or clearly defined features, facilitating the CNN's ability to extract significant patterns despite the removal of numerous weights. The model retains satisfactory accuracy with up to 70\% pruning and continues to learn rapidly. However, beyond this threshold, particularly at 90\%, the model begins to take considerably more time to converge, which may be due to the excessive loss of parameters affecting its learning capacity.

Additionally, Fig. \ref{fig:loss_all_images} reinforces these findings by illustrating the loss curves.  The observed loss trends correspond closely with the previously discussed accuracy patterns: for both Ton\_IoT and X-IIoTID (Fig. \ref{fig:loss1} and Fig. \ref{fig:loss2}), the loss remains consistently low and stable until moderate pruning levels are reached, after which it increases significantly once surpassing 60\%.  Conversely, the IDSIoT2024 dataset (Fig. \ref{fig:loss3}) maintains a relatively low loss even with 70\% pruning, which further substantiates its robustness against significant weight pruning.

Fig. \ref{fig:noniid_ACC_all_images} shows how pruning affects model accuracy, this time with Non-IID data using the FedAvg algorithm. Like before, we study how the models behave on the three IoT datasets: Ton\_IoT, IDSIoT2024, and X-IIoTID. We look at their performance over 40 communication rounds and different pruning levels, from from 0\% to 90\%.

The Ton\_IoT dataset, as illustrated in Fig. \ref{fig:noniid_1}, shows a strong sensitivity to both pruning and data heterogeneity. As  $\rho$  goes beyond 30\%, the model’s performance starts to drop noticeably, and its convergence becomes less stable, especially at higher pruning levels like 70\%, 80\%, and 90\%. On top of that, the non-IID nature of the data makes the training process even harder, since the global model tends to move away from its optimal weights after each round of aggregation.

As shown in Fig. \ref{fig:noniid_2}, the X-IIoTID dataset demonstrates greater resistance to pruning and data heterogeneity than Ton\_IoT. However, its performance still falls short compared to the IID-based model. From the figure, it is evident that when $\rho$ is below 40\%, the unpruned model continues to learn effectively, maintaining performance close to that of the original model. Beyond this threshold, however, convergence performance degrades as the pruning level increases.

Finally, as illustrated in Fig. \ref{fig:noniid_3}, the IDSIoT2024 dataset demonstrates greater robustness to both pruning and non-IID client data distribution compared to the other datasets. There is less performance degradation relative to IID data. This could be explained by the fact that this recent dataset contains features that are well-representative, allowing our CNN-based model to extract enough information for learning. As seen in the figure, for $\rho$ below 60\%, the performance remains similar to that of the unpruned model. However, beyond this threshold, the model starts to lose performance.

Fig. \ref{fig:noniid_ACC_all_images_fedprox} presents the evolution of model accuracy across 40 communication rounds under Non-IID data settings, using the FedProx algorithm. The results are shown for three IoT datasets:  Ton\_IoT, IDSIoT2024, and X-IIoTID, at various pruning levels ranging from 0\% to 90\%.

FedProx is used as an improvement over the standard FedAvg algorithm. It helps the global model train more effectively when the data is different across clients (non-IID). As shown in Fig.~\ref{fig:noniid_fedprox_img1}, Fig.~\ref{fig:noniid_fedprox_img2}, and Fig.~\ref{fig:noniid_fedprox_img3}, the three datasets show better training performance and are less affected by pruning when using FedProx instead of FedAvg. with FedProx, the models can handle pruning levels up to 50\% for Ton\_IoT, up to 60\% for X-IIoTID, and up to 70\% for IDSIoT2024, without losing much performance compared to the unpruned model. However, if the pruning rate goes beyond these levels, model performance drops significantly and training becomes unstable. This shows that FedProx helps maintain stable training in non-IID settings and improves the reliability of pruned models up to a certain limit.

Fig.~\ref{fig:noniid_loss_fedprox_all_images} confirms and strengthens the previous results, demonstrating that FedProx not only improves the loss convergence behavior in challenging Non-IID scenarios, but also enhances the model's robustness to moderate $\rho$. This is mainly due to the proximal regularization term in the FedProx loss function, which stabilizes local updates and prevents drastic weight changes, leading to  more stable convergence.

\begin{figure}[ht]
\centering
\includegraphics[width=0.8\columnwidth]{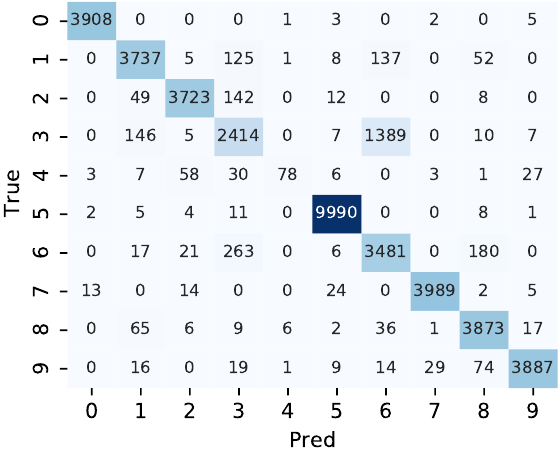}
    \caption{ Confusion matrix of our proposed CNN model  on \\ Ton\_IoT dataset under non-IID settings with FedProx.}
    \label{fig:ton_iot}
\end{figure}

\begin{figure}[ht] 
    \centering
\includegraphics[width=0.9\columnwidth]{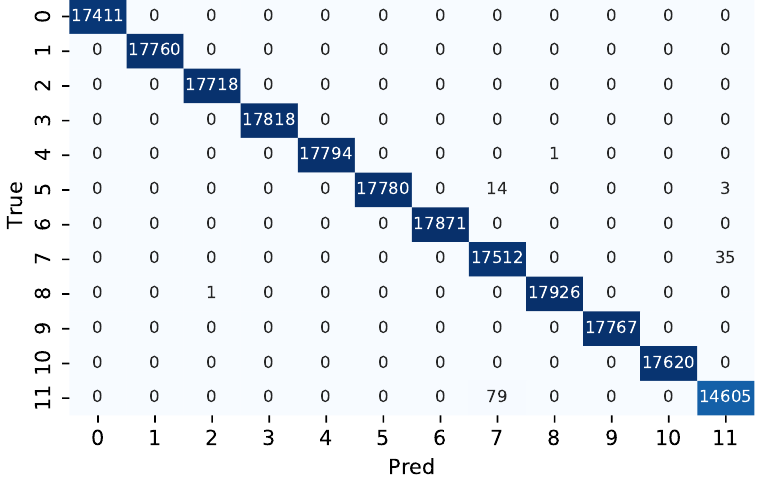}
    \caption{  Confusion matrix of our proposed CNN model on IDSIoT2024 dataset under non-IID settings with FedProx.}
    \label{fig:idsiot}
\end{figure}

\begin{figure}[ht]
\includegraphics[width=1\columnwidth]{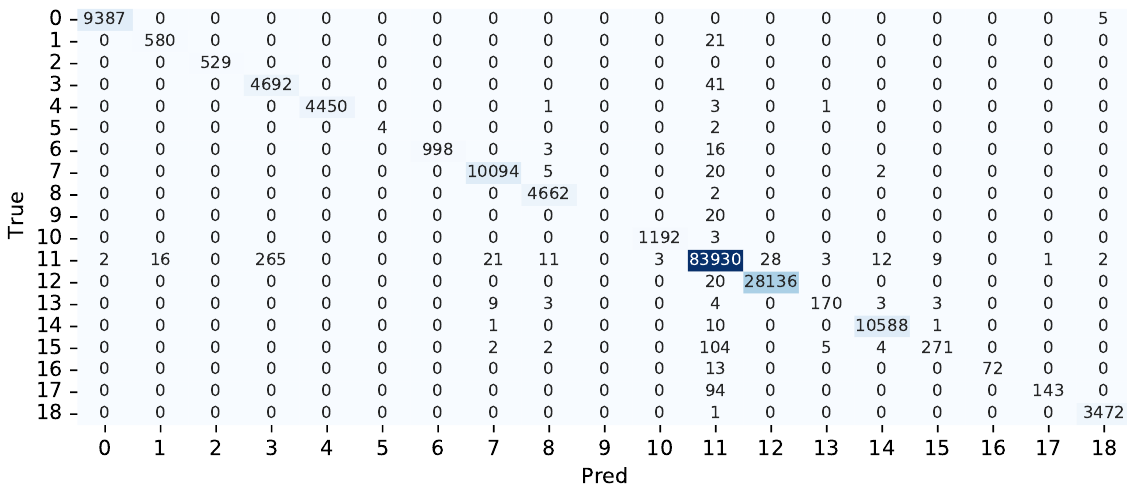}
     \caption{ Confusion matrix of our proposed CNN model on   X-IIoTID  dataset under non-IID settings with FedProx.}
    \label{fig:x_iiotid}
   
\end{figure}

Fig.~\ref{fig:ton_iot}, Fig.~\ref{fig:idsiot}, and Fig.~\ref{fig:x_iiotid} show the confusion matrices of our CNN model trained with FedProx under non-IID conditions on Ton\_IoT, IDSIoT2024, and X-IIoTID datasets. Each matrix compares true and predicted classes, numbered according to Table~\ref{tab:datasets}, where 0 corresponds to the first attack listed, 1 to the second, etc.

Fig.~\ref{fig:ton_iot}, for the Ton\_IoT dataset, shows that the model performs well on dominant classes, correctly classifying 9,990 normal traffic instances (5) and performing strongly on ransomware (7), backdoor (0), XSS (9), scanning (8), DDoS (1) and DoS (2). However, there is a notable confusion between injection (3) and password (6) attacks, with 1,389 injection samples misclassified as password attacks. This can be explained by similar feature patterns between these attacks and class imbalance in the dataset. These results highlight the model’s ability to recognize major categories effectively, while also exposing persistent challenges in distinguishing between attack types with similar behaviors.

Fig.~\ref{fig:idsiot} presents the results on the IDSIoT2024 dataset under non-IID conditions. The results demonstrate strong classification performance, with most predictions concentrated along the diagonal, indicating high accuracy across all classes. The matrix highlights the model's robustness in handling heterogeneous data distributions typical in  FL scenarios

As shown in Fig.~\ref{fig:x_iiotid}, the model performs strongly on the majority classes in the XIIoTID dataset, particularly Normal (11) with 83,930 correct predictions and RDoS (12) with 28,136, demonstrating effective learning on well-represented categories. However, it struggles with high minority classes, recording zero correct predictions for MITM Attacks (9) and only four for Generic Scanning (5), likely due to insufficient training data. Despite these limitations, the model maintains acceptable overall performance, largely driven by its success on the dominant classes.

The optimization problem reformulated in (\ref{eq:objectif_normalise}), which aims to determine the optimal value of $\rho$ that minimizes energy consumption during inference while maintaining acceptable model accuracy, is addressed using DRL. To this end, we define a score that serves as the objective function to be optimized. This score is designed to balance the trade-off between model accuracy and energy efficiency. It is computed using the accuracy of each local model trained locally on each client, evaluated after testing our FL system with FedProx without pruning, as shown in Table~\ref{tab:DRL}.  We set $\alpha_1 = 1$ to weight the accuracy term, and dataset-specific values of $\alpha_2$ are used to normalize the inverse of energy consumption: 50000 for the  Ton\_IoT dataset, 90500 for X-IIoTID, and 108000 for  IDSIoT2024. The energy consumption for each client is estimated based on the number of FLOPs and the model size in MB, with $B = 4$, $E_{\text{FLOP}} = 2.3$ pJ, and $E_{\text{access}} = 640$ pJ \cite{10398587}.

The DRL training is performed with $\lambda = 10$ and $\beta = 0.00002$. The agent is trained to solve the optimization problem by selecting the pruning rate $\rho$ that maximizes the defined score, achieving an optimal balance between energy reduction and performance preservation. As shown in Fig. \ref{fig:3}, the X-IIoTID dataset achieves the highest score with a pruning rate of $\rho = 68.36\%$ and a final score of 1.0325. The IDSIoT2024 dataset reaches a best score of 1.0323 with an optimal pruning rate of 65.85\%, while the  Ton\_IoT dataset achieves a best score of 0.9699 with its corresponding  best pruning rate 65.75\%. These results are consistent with the experimental findings presented earlier, demonstrating that our DRL-based method effectively solves the pruning rate optimization problem by enabling more than 60\% energy reduction compared to the unpruned model, while also decreasing model complexity with minimal accuracy degradation.

\section{CONCLUSION AND FUTURE WORK} \label{D}
In this paper, we introduce OptiFLIDS, a FL-based IDS framework in the context of critical and resource-constrained IoT environments. By integrating a one-time, non-progressive pruning technique applied at the initial training round and adapting the aggregation strategy using FedProx, we effectively reduced computational complexity and improved energy efficiency while maintaining high intrusion detection performance. Experimental results using three recent IoT datasets demonstrated the robustness and practicality of the proposed method. Our approach maintained high accuracy across heterogeneous client data, even with a pruning rate reaching up to 60\% in most cases. This resulted in a significant reduction in model parameters and thus improved energy efficiency. While promising, the framework has some limitations, particularly the initial communication overhead of pruning masks and the lack of evaluation at larger scale (e.g., 1000+ clients), which may challenge aggregation consistency and communication efficiency. We also aim to integrate proactive threat detection methods \cite{10945179,EchChammakhy2025EventHunterDC} to anticipate and mitigate potential intrusions before they occur, thereby strengthening the system’s resilience. Furthermore, we intend to improve the scalability of OptiFLIDS and to explore explainability techniques to enhance the  interpretability of model decisions, despite the added optimization complexity. In future work, we also plan to conduct a comparative study between our pruning-based approach and other model compression techniques, such as \textit{quantization} and \textit{knowledge distillation}, to better understand their trade-offs in terms of model performance and energy consumption within FL environments.

\bibliographystyle{unsrt} 

\bibliography{biblio}

\end{document}